\documentclass{article} %
\usepackage{iclr2025_conference,times}
\usepackage[T1]{fontenc}
\usepackage{minted}

\usepackage{amsmath,amsfonts,bm}

\def\eqref#1{equation~\ref{#1}}

\def\1{\bm{1}}

\DeclareMathAlphabet{\mathsfit}{\encodingdefault}{\sfdefault}{m}{sl}
\SetMathAlphabet{\mathsfit}{bold}{\encodingdefault}{\sfdefault}{bx}{n}

\usepackage{hyperref}
\usepackage{url}
\usepackage{xcolor,colortbl}
\usepackage{booktabs}
\usepackage{multirow}
\usepackage{algorithm}
\usepackage{algpseudocode}
\usepackage{todonotes}
\usepackage{enumitem}
\usepackage{subcaption}
\usepackage{float}
\newcommand{\benchmarkname}{\textsc{OpaqueToolsBench}\xspace}
\newcommand{\methodname}{\textsc{ToolObserver}\xspace}
\newcommand{\methodnameshort}{\textsc{TO}\xspace}
\usepackage{xspace}  %
\usepackage{graphicx}
\usepackage{cleveref}
\usepackage{tcolorbox}
\usepackage{tabularx}
\usepackage{listings}
\lstdefinestyle{prompt}{
    backgroundcolor=\color{gray!5},
    basicstyle=\ttfamily\small,
    breaklines=true,
    frame=single,
    xleftmargin=2pt,
    xrightmargin=2pt,
    framesep=2pt,
    aboveskip=10pt,
    belowskip=5pt,
    captionpos=b
}
\usepackage{array}
\usepackage{pgfplots}
\usetikzlibrary{calc}
\pgfplotsset{compat=1.18}
\usepackage{wrapfig}
\tcbuselibrary{skins, breakable, raster}
\usepackage{colortbl}
\usepackage[bottom]{footmisc}
\definecolor{tableheader}{RGB}{70, 130, 180}
\definecolor{tablerowlight}{RGB}{240, 248, 255}
\definecolor{tablerowalternate}{RGB}{255, 255, 255}

\definecolor{lightgray}{gray}{0.92}
\definecolor{darkergray}{gray}{0.75}  %
\newif\ifshowcomments

\newcommand{\jack}[1]{\ifshowcomments\todo[color=blue!40,linecolor=blue,size=\footnotesize]{#1 -JH}\fi}

\newcommand{\ashwin}[1]{\ifshowcomments\todo[color=violet!40,linecolor=violet,size=\footnotesize]{#1 -AP}\fi}

\title{\benchmarkname: Learning Nuances of Tool Behavior Through Interaction}

\author{
  Skyler Hallinan$^{\heartsuit *}$ \quad Thejas Venkatesh$^{\diamondsuit}$ \quad Xiang Ren$^{\heartsuit}$ \quad Sai Praneeth Karimireddy$^{\heartsuit}$ \\
  \textbf{Ashwin Paranjape}$^{\diamondsuit}$ \quad \textbf{Yuhao Zhang}$^{\diamondsuit}$ \quad \textbf{Jack Hessel}$^{\diamondsuit}$\\[0.25em]
  {\normalsize $^{\heartsuit}$University of Southern California \quad $^{\diamondsuit}$Samaya AI}\\[0.1em]
  {\texttt{skyler.r.hallinan@gmail.com}}
}

\iclrfinalcopy %
\begin{document}

\maketitle

\begin{abstract}
\begingroup
  \renewcommand\thefootnote{\fnsymbol{footnote}}
  \footnotetext[1]{Work done while interning at Samaya AI}
\endgroup
Tool-calling is essential
for Large Language Model (LLM) agents to complete real-world tasks.
While most existing benchmarks assume simple, perfectly documented tools, real-world tools (e.g., general ``search" APIs) are often \emph{opaque,} lacking clear best practices or failure modes. %
Can LLM agents improve their performance in environments with opaque tools by interacting and subsequently improving documentation?
To study this, we create \benchmarkname, a benchmark consisting of three distinct task-oriented environments: general function calling, interactive chess playing, and long-trajectory agentic search. 
Each environment provides underspecified tools that models must learn to use effectively to complete the task.
Results on \benchmarkname suggest existing methods for automatically documenting tools are expensive and unreliable when tools are opaque. %
To address this, we propose a simple %
framework, \methodname, that iteratively refines tool documentation by observing execution feedback from tool-calling trajectories.
Our approach %
outperforms existing methods on \benchmarkname across datasets, even in relatively hard settings. Furthermore, for test-time tool exploration settings, our method is also efficient, consuming 3.5-7.5$\times$ fewer total tokens than the best baseline. %
\footnote{We release our code and data at \url{https://github.com/shallinan1/OpaqueToolsBench}}

\end{abstract}

\section{Introduction}
\begin{figure}[t!]
    \vspace{-10pt}
    \centering
    \includegraphics[width=0.9\linewidth]{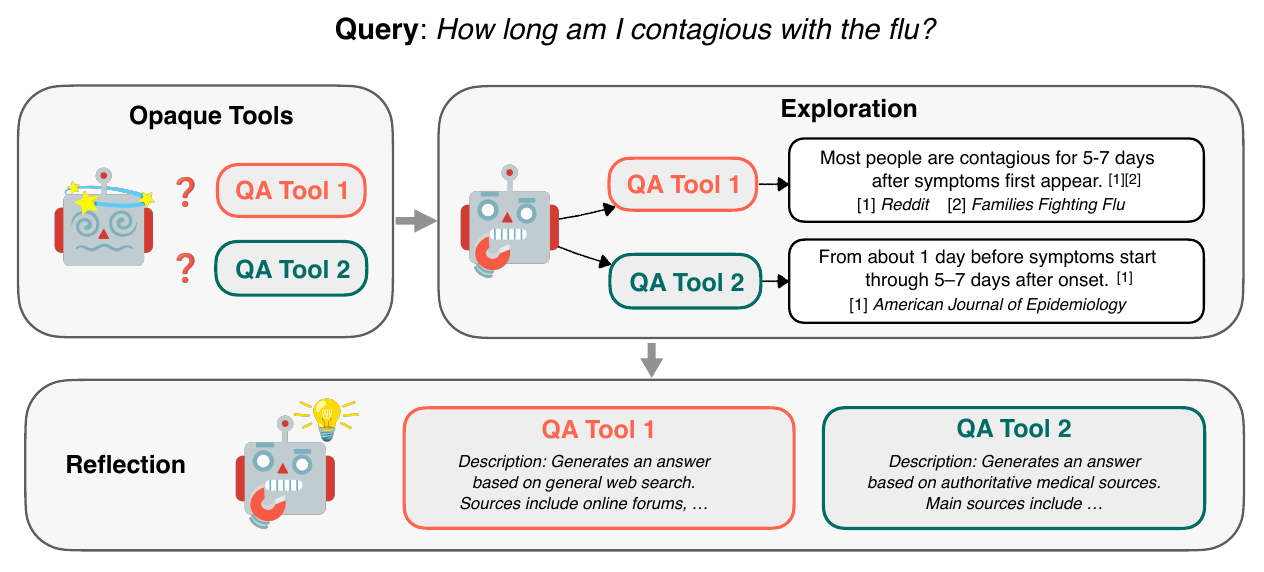}
    \vspace{-5pt}
    \caption{LLM agents may struggle when presented with \emph{opaque tools} -- tools lacking clear description of their usage best practices or their failure modes. To succeed in these settings, we posit that LLM agents must explore tool usage to learn their true behaviors.}
    \vspace{-5pt}
    \label{fig:placeholder}
\end{figure}

Tools expand the knowledge and capabilities of Large Language Model (LLM) agents beyond their learned parameters \citep{Schick2023ToolformerLM}.
With the right tools, LLMs can search the web, send emails, execute code, and interact with the world. Yet this extension fails when tools lack adequate documentation -- a pervasive problem in deployed systems. Real-world APIs are complicated to explain, enterprise functions lack specifications, and domain-specific tools ship with minimal descriptions. Moreover, some tools are difficult (or impossible) to fully explain, even for humans, e.g., often the optimal usage of search engines or LLM-based QA APIs is not known, even to their creators. Such tools are \textbf{opaque}: their behavior unpredictable, their correct usage unknown. As shown in \Cref{fig:placeholder}, tool opacity harms the performance of LLM agents, a problem that compounds when models must coordinate multiple tools for complex tasks.

This raises a fundamental question: \textbf{Can LLM agents learn to use opaque tools by observing their behavior during interaction?} Such capability would transform tool-calling agents from brittle systems dependent on perfect specifications into adaptive ones that can improve through experience. %

We introduce \benchmarkname, a benchmark for learning in opaque tool settings where tool documentation is underspecified. Different than existing benchmarks that assume near-perfect tool specifications like ToolBench \citep{Qin2023ToolLLMFL}, APIBench \citep{Patil2023GorillaLL}, and Berkeley Function Calling Leaderboard \citep{patil2025bfcl}, \benchmarkname does \emph{not} provide comprehensive function signatures, detailed descriptions, or explicit type specifications. \benchmarkname spans three distinct environments: general function calling, interactive chess playing, and long-trajectory agentic search. Each provides underspecified tools that models must learn to use effectively, testing both single-instance discovery and cross-instance learning. Unlike existing benchmarks that focus on measuring model capacity to compose concrete, well documented tools, \benchmarkname is designed to measure an LLM agent's ability to adapt to the imperfect documentation of tools through interaction.

We evaluate current approaches that optimize tool descriptions, including Play2Prompt \citep{Fang2025PLAY2PROMPTZT} and EasyTool \citep{Yuan2024EASYTOOLEL}. Both methods fall short on \benchmarkname:  systems either focus on compression of existing tool documentation, completely neglecting the interaction with tools (EasyTool) or require single-tool exploration phases separate from task execution, which becomes expensive in some settings (Play2Prompt).
As a result, these methods are ineffective and sometimes expensive, often consuming thousands of tokens in preliminary exploration before attempting the composite task. %

We propose an alternative framework, \methodname, that refines tool documentation by observing and learning from execution feedback acquired through composite task trajectories. %
On \benchmarkname, our method exceeds baseline performance consistently by on average 18.6\%, while consuming 3.5-7.5× fewer tokens in test-time settings.
Our results demonstrate that learning from execution feedback provides an efficient path to handling opaque tools in real-world environments.
It also demonstrates that LLM agents can adapt to underspecified tools through interaction, making tool-calling viable even in poorly-documented environments.
\section{Background}

Language models estimate the conditional distribution $P(x_t | x_{<t})$ over a vocabulary $\mathcal{V}$, where $x_{<t}$ represents all preceding tokens \citep{Radford2019LanguageMA, Brown2020LanguageMA}. 
Despite their impressive results on language benchmarks, language models nonetheless face fundamental limitations: their knowledge is frozen, they can't take actions in the real world, and they are often unreliable on simple capabilities like adding numbers. %
These constraints have motivated the development of \emph{tool-augmented} language models that invoke external functions to overcome these limitations \citep{Schick2023ToolformerLM, Mialon2023AugmentedLM}.

\textbf{Tool-Calling in Language Models} 
Language models interact with tools through a structured interface that enables them to extend their capabilities beyond parametric knowledge. Each tool $t_i$ in the available tool library $\mathcal{T} = \{t_1, t_2, \ldots, t_n\}$ is characterized by:
\begin{align*}
t_i = \{n_i, d_i, p_i, e_i\}
\end{align*}
where $n_i$ is the name of the function, $d_i$ is the documentation of the function behavior, $p_i$ is an optional dictionary of parameters consisting of their name, description, and whether or not they are required.  $e_i$ is the executable function itself. In practice, parameter information from
$p_i$ is often incorporated into the behavioral documentation $d_i$ as a string description.

Given a user query $q$, the language model $M$ generates tool calls through a systematic process of \textbf{retrieving} then \textbf{calling}, all through autoregressive decoding. The model conditions on both the query and the available tool descriptions in their context to produce a sequence that may include tool invocations:
\begin{align}
s \sim P_M(s | q, \mathcal{T}) = P_M(s | q, {(n_1, d_1, p_1), \ldots, (n_n, d_n, p_n)})
\end{align}
where $s$ is the sampled sequence. When the model determines a tool is needed, it produces a structured tool call $c = \langle n_i, \text{args}_i \rangle$ as part of this sequence, where $\text{args}_i$ must conform to the parameter specification $p_i$.
Upon generating a tool call, the system executes $r_i = e_i(\text{args}_i)$ and appends the result to the context. The model then continues its autoregressive generation, now conditioning on the expanded context:
\begin{align}
s' \sim P_M(s' | q, \mathcal{T}, c, r_i)
\end{align}
This process may repeat, with the model invoking multiple tools or generating a final response that incorporates the tool outputs to address the user's query. Such tool calling language models are often used to complete goal-oriented tasks and are referred to as LLM agents \citep{Wang2023ASO}.

\section{The \benchmarkname Benchmark}

Prior work creating and evaluating LLM agents assumes well-defined and well-documented tools \citep{guo-etal-2024-stabletoolbench, Qin2023ToolLLMFL, Shen2024ShortcutsBenchAL, Patil2023GorillaLL}. However, many real-world tools are black boxes whose behavior can only be understood through interaction. As an example,
consider an agent given access to a set of off-the-shelf Search APIs provided as tools. Such APIs may index documents at differing granularities and covering different domains/times/topics/styles, may (or may not) be keyword based, might automatically run multiple hops, and could even themselves use language models to orchestrate the search process. %
For LLM agents to perform optimally in such an environment, they need to learn these nuances by \textit{interacting with these tools} and \textit{observing the feedback}.
Furthermore, the proliferation of Model Context Protocols (MCP) \footnote{modelcontextprotocol.io} have made it easy to connect LLM agents to tools that come with variable quality and accuracy of tool documentation. 

We characterize such tools as \textbf{opaque}. Reflecting on the diverse challenges described above, from the inherent complexity of Search APIs to the variable documentation quality of MCP tools, we find that opacity stems from two distinct sources. We distinguish between two types of opacity -- both practically important -- that motivate our benchmark design:

\begin{itemize}[leftmargin=0.5cm,]
    \item \textbf{Type 1: Documentation Opacity} Tools whose behavior is deterministic and describable, but whose documentation is inaccurate or missing. This is often inevitable in real-world settings: legacy systems often rely on unwritten ``tribal'' knowledge, while open marketplaces like the Model Context Protocol (MCP) contain inconsistent descriptions at scale.

    \item \textbf{Type 2: Intrinsic Opacity} Tools that are opaque due to inherent complexity (e.g., search engines, LLM-based tools, complex simulations). These tools often have a simple schema but complex, undocumented behavioral nuances. For example, a search API’s ranking logic is hard to fully capture a priori; even the tool creator cannot predict how a neural system handles every edge case. The agent must instead learn these nuances through interaction.
\end{itemize}

To effectively operate in environments characterized by these forms of opacity, LLM agents need to demonstrate the following abilities:
\begin{enumerate}[leftmargin=1cm]
\item \textbf{Manipulate structured \& natural language inputs:} Certain tools that expose traditional REST APIs (for example, currency conversion) have structured inputs. Whereas others like Search APIs accept a string where the nuances are encoded in natural language, which are more open-ended/opaque. 
\item \textbf{Learn from process feedback:} Opaque tools can sometimes only be understood by observing a trajectory/sequence of tool uses, rather than just the result from a single call, e.g., you might not be able to discern if a search API's output is ``good" until you try to compose the result with other information.
\item \textbf{Learn across trajectories:} In many real-world settings with a fixed set of tools, LLM agents need to accumulate experience from prior trajectories, e.g., using a search tool with one goal in mind should help one gain experience useful for using that tool for a different goal. %
\item \textbf{Test-time generalization:} In settings where new tools are available at test time, we need to test the ability of LLM agents to learn their nuances while completing the task itself. 
\item \textbf{Learn tool sequencing:} Certain tool behaviors may only manifest in conjunction with other specific tool calls. For example, if tool B can only be called after a successful invocation of tool A, the agent needs to be able to create such trajectories to learn nuances of tool B. %
\end{enumerate}

\begin{figure}[t!]
    \centering
\vspace{-10pt}
\includegraphics[width=0.99\linewidth, trim=0 130 0 0, clip]{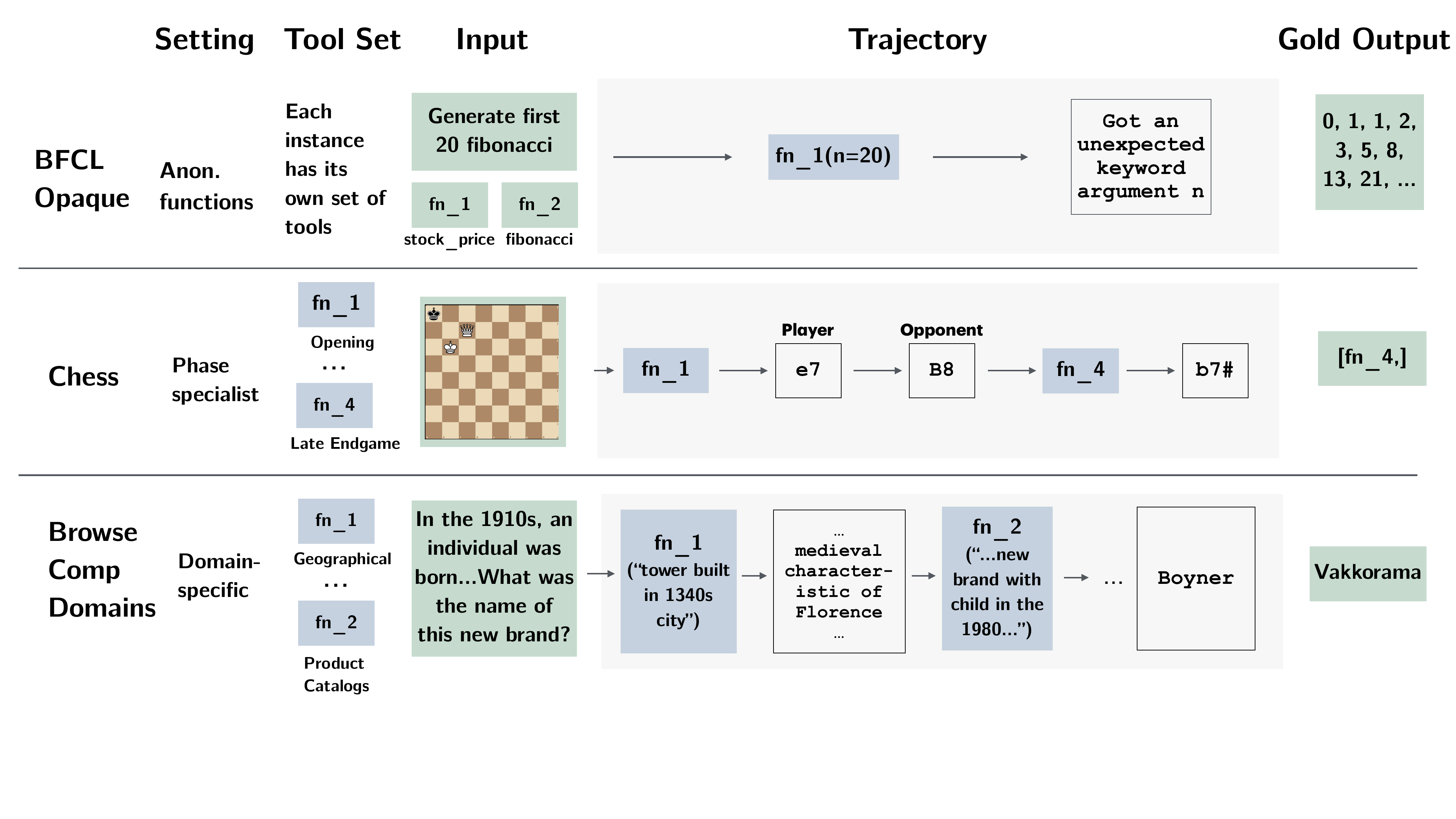}
\vspace{-10pt}
\caption{Examples from each of the environments in \benchmarkname. For Chess and BrowseComp Domains, each setting has a fixed set of tools across all instances (chess and search engines respectively). In BFCL-Opaque, each instance has custom query-dependent tools. The dataset defines the input and output (green). For each instance, the agent makes tool calls iteratively (blue). For BFCL-Opaque and BrowseComp Domains we check for a match with the gold answer. For Chess, we match engine choices in green with optimal engine choices. }%
\label{fig:benchmark_examples}
\end{figure}

We introduce \benchmarkname to systematically evaluate these abilities across both types of opacity. It consists of three environments: an opacified version of BFCL \citep{Patil2023GorillaLL} we call BFCL-Opaque (targeting \textbf{Type 1}), a novel game-playing environment based on Chess (targeting \textbf{Type 2}), and BrowseComp Plus with opaque search \citep{Chen2025BrowseCompPlusAM} (targeting \textbf{Type 2}).
Our environments are designed to test the above aspects of learning the behavior of opaque tools while being reproducible and efficient to run. A summary of our datasets and key information is shown in \Cref{tab:benchmark_overview_transposed} and in the following paragraphs. Examples from the tasks are shown in \Cref{fig:benchmark_examples}.

\begin{table}[t!]
\footnotesize
\centering
\caption{An overview of the datasets in \benchmarkname}
\vspace{-8pt}
\label{tab:benchmark_overview_transposed}
\
\begin{tabularx}{\textwidth}{
    l
    >{\raggedright\arraybackslash}p{0.25\textwidth}
    >{\raggedright\arraybackslash}p{0.25\textwidth}
    >{\raggedright\arraybackslash}p{0.25\textwidth}
}
\toprule
 & \textbf{BFCL-Opaque} & \textbf{Chess} & \textbf{BrowseComp Domains} \\
\midrule
\textbf{Description} & Complete user's request by correctly calling a function with a limited description.  & Play chess by choosing one of many specialized chess engine tools every turn. & Compose domain-specific search tools to locate hard-to-find information.  \\
\addlinespace
\textbf{Skills required} & 
Structured inputs, process feedback, test-time generalization  &
Process feedback, learn across trajectories, tool sequencing
& Unstructured inputs, process feedback, learn across trajectories, tool sequencing
\\
\textbf{Settings} & 
\textbf{Documentation quality}
\begin{enumerate}[leftmargin=*,after=\vspace{-0.6\baselineskip},itemsep=0pt,parsep=0pt,topsep=0pt]
    \item Anon. function names
    \item Anon. function names + description
    \item Anon. function names + parameter names
\end{enumerate} & 
\textbf{Tool sets (chess engines)}
\begin{enumerate}[leftmargin=*,after=\vspace{-0.6\baselineskip},itemsep=0pt,parsep=0pt,topsep=0pt]
    \item Beginner, intermediate, advanced skill
    \item Opening, midgame, endgame, late-endgame specialists
\end{enumerate}  & 
\textbf{Tool sets (search tools)}
\begin{enumerate}[leftmargin=*,after=\vspace{-0.6\baselineskip},itemsep=0pt,parsep=0pt,topsep=0pt]
    \item Domain-specific (9)
    \item Domain-specific (9) + Full Search
\end{enumerate}  \\
\textbf{Evaluation} & Evaluation Accuracy, Param Acc., AST Acc. & \% of Optimal Tool Calls, ELO & Accuracy, \# Tool Calls \\
\addlinespace
\textbf{\# Train / \# Test} & -  / 150 & 200  / 1800 & 83 / 747 \\
\bottomrule
\end{tabularx}
\end{table}

\textbf{BFCL-Opaque:} The Berkeley Function Calling Leaderboard (BFCL) \citep{patil2025bfcl} consists of question-functions-answer tuples with functions from multiple programming languages and diverse application domains. Following \citealp{Fang2025PLAY2PROMPTZT}, we use the executable subset so feedback is available from executing the tools. The tasks are simple, for example, fetching the current weather or computing the area of a polygon. Each task has between 2-4 tools and on average 3.
We modify this environment to evaluate \textbf{test-time generalization} and \textbf{Type 1 Opacity} -- though the tools of each instance may slightly overlap a few others instances', we instead treat the tools of each test instance as independent and unknown. Specifically, we modify this environment by obfuscating the function names as well as completely removing all docstrings. We do this independently for each problem -- so the same function across different test instances will be named differently. For each test instance, the agent now has to learn what the function does and produce structured output in the form of its arguments and their types. 
We create progressively easier settings by providing either (1) only function description or (2) only parameter names. We use a binary task completion rate as our evaluation metric.
Refer to \Cref{appendix:BFCL} for details on environment construction. 

\textbf{Chess} In this environment %
the task is to maximize playing strength against a series of increasingly hard opponents (Stockfish \citep{stockfish} configured with ELO ratings 1200, 1800, and 2400). Instead of playing moves directly (as in \cite{Zhang2025CompleteCG}), models are given access to a set of move suggesting tools. At each turn, the LLM agent uses one of several undocumented tools that accept current board positions in FEN notation and play the move that's suggested. While the tool interfaces are identical, we introduce undocumented behaviors in each tool. Specifically, in the first setting, each tool uses Stockfish configured with different ELO ratings (1200, 1800 \& 2400), testing if the agent can learn fine-grained discrimination between tools based on the final outcome. In the second setting, we provide tools optimized for opening, middlegame, endgame, and late endgame, enabling us to test the ability to explore and learn temporal patterns. 
Since the same set of tools are used across different instances, we test agent's ability to learn tool behavior across trajectories (capturing \textbf{Type 2 Opacity} where behavior depends on game state).
Refer to \Cref{appendix:Chess} for details on environment construction.

\textbf{BrowseComp Domains} 
The BrowseComp dataset consists of question and short answer pairs which measure the ability of agents to locate hard-to-find, entangled information on the internet, and might require browsing tens or even hundreds of websites in the process. BrowseComp Plus \citep{Chen2025BrowseCompPlusAM} further improves it by fixing the retrieval corpus and thus making this benchmark easy to reproduce. We leverage this corpus and create anonymous domain-specific search tools (4-6 tools) by partitioning the corpus (academic papers, product catalogs, geographical data, news articles). 
The LLM agent must not only discover the domain specialization of each tool, it must also learn to create optimal search queries for each tool and learn to search in the right sequence (addressing \textbf{Type 2 Opacity} in query formulation). Each instance in our dataset has a corresponding sequence of optimal tool calls. We compare the tool sequence with the optimal sequence and use accuracy as our evaluation metric. Refer to \Cref{appendix:BrowseComp} for details on environment construction. 

These three environments test for the four key abilities of LLM agents: (1) With BFCL-Opaque we test for structured inputs and with BrowseComp Domains unstructured inputs, (2) All environments allow learning from intermediate outputs of tool calls, (3) BrowseComp Domains and Chess test if the LLM agent can learn across trajectories, while BFCL-Opaque checks for generalization to new tools at test time, (4) Chess and BrowseComp Domains test for LLM agent ability to sequence opaque tools in the right sequence. 

\section{\methodname: Improving Documentation via Interaction}

\vspace{-0.6em}
Existing approaches for optimizing tool documentation have limitations that render them less effective in complex tool-calling environments.
For example, EasyTool~\citep{Yuan2024EASYTOOLEL} relies primarily on compressing existing long documentation of tools into more concise instructions, making it unsuitable to settings where the tool documentation is underspecified or completely lacking.
Meanwhile, Play2Prompt~\citep{Fang2025PLAY2PROMPTZT} adopts an iterative refinement strategy, but it requires an isolated single-tool exploration phase separate from task execution, making it inefficient and (as we will show) not performant in complex environments that require a long trajectory of tool calls.

To address these limitations, we propose \methodname. %
\methodname requires no up-front documentation, and discovers and refines tool documentation through observation and reflection on execution trajectories. 
The key idea is to alternate between an \emph{Exploration Phase} and a \emph{Reflection Phase}. During exploration (line 3 of Algorithm~\ref{alg:main}), we collect execution trajectories using the current documentation $\mathcal{D}^{(k-1)}$, which can initially be null. There a separate reasoning model analyzes the trajectories, identifies patterns, and updates tool descriptions accordingly.  We repeat this process iteratively, improving tool descriptions and exploring better trajectories in the next iteration. 

\benchmarkname contains two settings: (1) with \textbf{shared tools} between train and test set and (2) with \textbf{unseen tools} that can only be observed at test time (and there is no train set). We instantiate \methodname in two modes to accommodate the two settings: \textbf{offline mode}, which pre-optimizes documentation during training when all test instances share a common tool set and  
\textbf{online mode}, which optimizes documentation at test time for previously unseen tools.

\subsection{Offline Mode}

\begin{wrapfigure}{r}{0.55\textwidth}
\vspace{-2.4em}  %
\begin{minipage}{\linewidth}
\begin{algorithm}[H]
\caption{\methodname: Offline mode}
\label{alg:main}
\begin{algorithmic}[1]
\Require Initial descriptions $\mathcal{D}_0$, iterations $K$, LLM Agent $M_A$, Editor LLM $M_E$
\Ensure Optimized documentation $\mathcal{D}^*$
\State Initialize $\mathcal{D}^{(0)} \leftarrow \mathcal{D}_0$
\For{$k = 1$ to $K$}
    \State \textbf{// Exploration Phase}
    \State $\mathcal{T}_k \leftarrow \text{CollectTrajectories}(\mathcal{D}^{(k-1)}, M_A)$ 
    
    \State \textbf{// Reflection Phase}
    \State $\mathcal{D}^{(k)} \leftarrow \text{ReflectAndUpdate}(\mathcal{T}_k, \mathcal{D}^{(k-1)}, M_E)$
\EndFor
\State \Return $\mathcal{D}^{(K)}$
\end{algorithmic}
\end{algorithm}
\end{minipage}
\vspace{-1.3em}  %
\end{wrapfigure}

In offline mode, we pre-optimize documentation using a set of training instances sharing the same tools. Algorithm~\ref{alg:main} outlines the procedure. The core intuition is to establish a feedback loop where the agent attempts tasks, observes execution outcomes against ground truth, and iteratively refines its understanding of tool behaviors from this signal. This approach leverages two key advantages of the offline setting: (1) multiple training instances and (2) the ability to evaluate the final answer. We discuss the two key phases below:

\textbf{Exploration Phase: Collecting Trajectories} 
The goal of this phase is to generate diverse interaction traces that reveal the latent behaviors and failure modes of the opaque tools.
To do this, we execute the agent $M_A$ on the training set $\mathcal{X}_{train}$ using the current tool documentation $\mathcal{D}^{(k-1)}$. By running across the entire training set, we ensure the agent explores tool behaviors across a wide variety of contexts (e.g., diverse chess positions or search queries) and diverse usage patterns. We use temperature sampling to collect a breadth of trajectories $\mathcal{T}_k = \{\tau_1, ..., \tau_{|\mathcal{X}_{train}|}\}$, where each trajectory $\tau_i$ captures the full sequential decision process: starting from the initial state, it records the alternating sequence of reasoning steps, tool calls, and environmental observations (e.g., tool execution outputs) leading to the final outcome.

\textbf{Reflection Phase: Reflecting and Updating Documentation} 
The goal of this phase is to distill raw interaction experience into explicit, generalizable tool documentation. However, processing the full set of trajectories $\mathcal{T}_k$ simultaneously is infeasible due to context window constraints and the difficulty of extracting consistent patterns from massive, noisy data streams. To address this, we employ a meta-prompting strategy with an \textbf{``Editor'' LLM $M_E$} (this can be the same as our LLM agent $M_A$). It analyzes trajectories via a hierarchical process:
\begin{enumerate}[leftmargin=0.5cm,topsep=0pt, itemsep=0pt]
    \item \textbf{Batch Analysis (with ground truth):} We first split the trajectories into mini-batches to manage context limits. For each batch, we explicitly task the Editor with updating the tool descriptions based on the execution history. Acting as a local reasoner, it identifies causal links between tool usage patterns and success/failure outcomes. Crucially, it is provided with the \textbf{ground truth} (e.g., the gold answer) or a \textbf{task performance signal} derived from the training environment. Using this signal, it distinguishes effective usage from failures and generates a candidate description specific to that batch. Consequently, this phase produces a diverse set of local descriptions, where each captures insights valid for its specific subset of trajectories.
    \item \textbf{Consensus Merge:} Since observations from a single batch may be noisy or overfit to specific instances, a second Editor pass aggregates the candidate proposals from all batches. It acts as a \textbf{consensus filter}, retaining only those behavioral rules that appear consistently across multiple diverse batches while discarding instance-specific hallucinations. 
\end{enumerate}

This two-stage process enables accurate tool documentation, even with hundreds of long trajectories. We repeat this exploration and reflection loop for $K$ iterations. Finally, we run the LLM agent with the optimized documentation $\mathcal{D}^{(K)}$ on the test set. Further implementation details, including prompt templates and dataset-specific details, can be found in \Cref{appendix:experiments} and \ref{appendix:prompts}.

\subsection{Online mode}
In the online mode, we are only given access to a single test instance with a set of unseen tools. 
For a single test instance with unique tools $\mathcal{T}_x$ that cannot be experimented with apriori, we first collect a trajectory by executing the agent's generated tool calls and recording the real-time environment feedback (e.g., return values or error messages). Then, we use the \textbf{Editor LLM} to analyze that trajectory and update the tool documentation based on the observed behavior(s). 
We repeat this process for a maximum of $K$ iterations. If the Editor does not update the documentation, we stop the iteration early. 
Note that in online mode the gold output is not used as part of the process. 
We use the final version of the tool documentation to run the LLM agent on the test instance and compute an evaluation metric using the gold output.

\subsection{Comparison to Prior Interactive Methods}
Prior works like Play2Prompt \citep{Fang2025PLAY2PROMPTZT} and DRAFT \citep{Qu2024FromET} rely on ``isolated unit testing,'' generating synthetic queries to test tools in a vacuum. This approach misses sequential dependencies critical for tasks like chess. In contrast, our method employs \textit{trajectory-based learning}, optimizing documentation within actual multi-step tasks to capture inter-tool dynamics and opaque syntax. Furthermore, while prior methods require exhaustive \textit{a priori} optimization of the entire tool library, our method optimizes documentation \textit{on-demand}, focusing exploration only on the relevant subset of tools required by the current trajectory.

\section{Experiments}

 We benchmark three contemporary tool-calling LLM agents on \benchmarkname. First, we test on GPT-5 \citep{openai2025introducinggpt5}, the most capable model OpenAI model for reasoning and agentic tool use at the time of writing. We also use GPT-5-mini, a cost-efficient yet still capable version. We use the ReAct framework \citep{Yao2022ReActSR} to iteratively reason and call functions. Further details can be found in \Cref{appendix:experiments}.

\textbf{\methodname and Baselines Experimental Details}
For our main experiments with \methodname, we also use GPT-5 as our model for reflecting on and updating tool descriptions (editor model). F ollowing \methodname, for all baselines we use GPT-5. We include (1) \textbf{Play2Prompt} \citep{Fang2025PLAY2PROMPTZT}, which improves tool-documentation from self-play followed by self-reflection and (2) \textbf{EasyTool} \citep{Yuan2024EASYTOOLEL} which automatically rewrite the tool documentation by condensing tool descriptions and creating structured functional guidelines. 

  \begin{table}[t!]                                      
  \centering                                                                                                                         
  \caption{Performance on BFCL across models and baselines. We use ReAct. Gold is the ground truth documentation and base is the     
  opacified set. \methodnameshort denotes \methodname, P2P denotes Play2Prompt, and ET denotes EasyTool. Columns denote \textbf{E}
  (Execution-based overall accuracy), \textbf{P} (Parameter accuracy), and \textbf{A} (AST accuracy). The highest \textbf{E} value in
   each row is \textbf{bolded}.}
  \vspace{-5pt}
  \setlength{\tabcolsep}{3pt}
  \resizebox{\textwidth}{!}{
  \begin{tabular}{
      >{\raggedright\arraybackslash}p{2.2cm}
      c
      >{\columncolor{gray!10}}c
      >{\columncolor{gray!10}}c
      >{\columncolor{gray!10}}c
      c c c
      >{\columncolor{gray!10}}c
      >{\columncolor{gray!10}}c
      >{\columncolor{gray!10}}c
      c c c
      >{\columncolor{gray!10}}c
      >{\columncolor{gray!10}}c
      >{\columncolor{gray!10}}c
  }
  \toprule
  \multirow{3}{*}{\textbf{Documentation}} & \multirow{3}{*}{\textbf{Model}} & \multicolumn{15}{c}{\textbf{ReAct}} \\
  \cmidrule(l){3-17}
   & & \multicolumn{3}{c}{\cellcolor{gray!10}\textbf{Gold}} & \multicolumn{3}{c}{\textbf{Base}} &
  \multicolumn{3}{c}{\cellcolor{gray!10}\textbf{+  \methodnameshort}} & \multicolumn{3}{c}{\textbf{+ P2P}} &
  \multicolumn{3}{c}{\cellcolor{gray!10}\textbf{+ ET}} \\
  \cmidrule(lr){3-5} \cmidrule(lr){6-8} \cmidrule(lr){9-11} \cmidrule(lr){12-14} \cmidrule(l){15-17}
   & & \textbf{E} & \textbf{P} & \textbf{A} & \textbf{E} & \textbf{P} & \textbf{A} & \textbf{E} & \textbf{P} & \textbf{A} &
  \textbf{E} & \textbf{P} & \textbf{A} & \textbf{E} & \textbf{P} & \textbf{A} \\
  \midrule
  \rowcolor{gray!20}
  \multicolumn{17}{c}{Tool Setting: Individual Tools per Problem}\\
  \midrule

  \multirow{2}{=}{Anon. Fn. Names Only}
   & GPT-5
   & 0.95 & 0.96 & 0.96
   & 0.00 & 0.00 & 0.60
   & \textbf{0.80} & 0.78 & 0.89
   & 0.44 & 0.45 & 0.76
   & 0.00 & 0.00 & 0.60 \\

   & GPT-5-mini
   & 0.93 & 0.94 & 0.96
   & 0.00 & 0.00 & 0.60
   & \textbf{0.66} & 0.63 & 0.85
   & 0.43 & 0.43 & 0.77
   & 0.00 & 0.00 & 0.60 \\
  \midrule

  \multirow{2}{=}{Anon. Fn. + Real Desc}
   & GPT-5
   & 0.95 & 0.96 & 0.96
   & 0.00 & 0.00 & 0.60
   & \textbf{0.86} & 0.82 & 0.91
   & 0.53 & 0.54 & 0.82
   & 0.00 & 0.00 & 0.60 \\

   & GPT-5-mini
   & 0.93 & 0.94 & 0.96
   & 0.00 & 0.00 & 0.60
   & \textbf{0.84} & 0.81 & 0.91
   & 0.55 & 0.55 & 0.82
   & 0.00 & 0.00 & 0.60 \\
  \midrule

  \multirow{2}{=}{Anon. Fn. + Param Names}
   & GPT-5
   & 0.95 & 0.96 & 0.96
   & 0.82 & 0.85 & 0.96
   & \textbf{0.83} & 0.86 & 0.95
   & 0.79 & 0.81 & 0.95
   & 0.75 & 0.77 & 0.94 \\

   & GPT-5-mini
   & 0.93 & 0.94 & 0.96
   & 0.84 & 0.88 & 0.96
   & \textbf{0.85} & 0.89 & 0.96
   & 0.77 & 0.80 & 0.95
   & 0.76 & 0.79 & 0.95 \\

  \bottomrule
  \end{tabular}
  }
  \label{tab:bfcl_full}
  \end{table} 
\subsection{Main Results}
Our main results in Tables \ref{tab:bfcl_full}-\ref{tab:browsecomp_full} demonstrate both the challenge of existing baselines on \benchmarkname, and the strong performance of \methodname, showing its potential to improve LLM agent performance in opaque tool settings.

\textbf{\methodname outperforms baselines on BFCL-Opaque and recovers near-optimal performance.} \Cref{tab:bfcl_full} demonstrates that \methodname outperforms all baselines across documentation levels and LLM agents. In the hardest setting (Function Names Only), \methodname recovers \textbf{0.80} execution accuracy with GPT-5, whereas Play2Prompt only achieves 0.44. Furthermore, because Play2Prompt exhaustively tests hundreds of tools, most of which will not be relevant, the total input + output tokens consumed for exploration is $\sim$1.7M, which is 7.5$\times$ more than the exploration budget of \methodname. Across documentation levels, this ranges from 3.5$\times$ -- 7.5$\times$. This indicates that for test-time optimization settings, our method is both effective and relatively inexpensive.

We identify an initially challenging but tractable scenario: when parameter information is missing. In the ``Anon. Fn. + Real Desc'' setting, OpenAI agents initially obtain 0 performance, repeatedly calling tools without arguments. However, \methodname effectively bridges this gap. By comparing against a Gold Oracle (perfect documentation), we find our method recovers \textbf{90\%} of the performance gap in this underspecified setting (0.86 vs. 0.95). To verify this is due to valid schema discovery rather than luck, we analyze Parameter (P) and AST (A) accuracy. \methodname consistently outperforms baselines on these granular metrics (e.g., \textbf{0.78} Parameter Accuracy vs. 0.45 for Play2Prompt in the Anon. Names setting), confirming it successfully synthesizes correct input structures from scratch.

\begin{table}[t]
\footnotesize
\centering
\caption{Average percentage of best tool calls (Acc) and ELO rating at a given state for Chess. \methodnameshort denotes \methodname, P2P denotes Play2Prompt, and ET denotes EasyTool. Gold is the ground truth documentation. The highest value in each row is \textbf{bolded}.}
\vspace{-5pt}
 \begin{tabular}{lcccccccccc}
\toprule
\multirow{3}{*}{\textbf{Model}} & \multicolumn{10}{c}{\textbf{ReAct}} \\
\cmidrule(l){2-11}
 & \multicolumn{2}{c}{\textbf{Gold}} & \multicolumn{2}{c}{\textbf{Base}} & \multicolumn{2}{c}{\textbf{+ \methodnameshort}} & \multicolumn{2}{c}{\textbf{+ P2P}} & \multicolumn{2}{c}{\textbf{+ ET}} \\
\cmidrule(lr){2-3} \cmidrule(lr){4-5} \cmidrule(lr){6-7} \cmidrule(lr){8-9} \cmidrule(l){10-11}
 & \textbf{Acc} & \textbf{ELO} & \textbf{Acc} & \textbf{ELO} & \textbf{Acc} & \textbf{ELO} & \textbf{Acc} & \textbf{ELO} & \textbf{Acc} & \textbf{ELO} \\
\midrule
\rowcolor{gray!20}
\multicolumn{11}{c}{Tool Setting: Opening, midgame, endgame, late-endgame specialists} \\
GPT-5 & 64.4 & 1411  & 23.5 & 728 & \textbf{40.1} & 1020 & 35.8 & 966 & 23.2 &  761 \\
GPT-5-mini & 52.8 & 1243 & 24.9 & 772 & \textbf{32.1} & 949 & 19.5 & 754 & 25.8 & 739 \\

\midrule
\rowcolor{gray!20}
\multicolumn{11}{c}{Tool Setting: Beginner, intermediate, advanced skill} \\
GPT-5 & 100 & 2341  & 24.9 & 1572 & \textbf{29.1} & 1756 & 0.25 & 1622 & 24.9 & 1584 \\
GPT-5-mini & 100 & 2346 & 25.7 & 1674 & \textbf{28.4} & 1778 & 22.7 & 1481 & 25.5 & 1645 \\
\bottomrule
\end{tabular}
\label{tab:chess_full}
\end{table} 

\textbf{BrowseComp Domains and Chess are challenging, but \methodname can still discover tool nuances}
As shown in \Cref{tab:chess_full} and \ref{tab:browsecomp_full}, BrowseComp Domains and Chess are empirically much harder tasks than BFCL Opaque. For example, even with perfect documentation (Gold Oracle), GPT-5 only achieves 27.8\% accuracy on BrowseComp (Domain + Full Search). Despite this high difficulty, \methodname outperforms baselines in the majority of settings.

In the Chess domain, the absolute accuracy on the playing strength setting appears low (e.g., 29.1\% for GPT-5). We posit that this is because our "Best Tool \%" metric is strict; it penalizes the model for choosing the 1800 ELO tool over the 2400 ELO tool, even though both may suffice to beat a weaker opponent. However, the \textbf{Streaming ELO} scores---computed from actual win-rates against diverse opponents---verify that \methodname enables practically effective gameplay. By refining the tool descriptions to reflect strategic capabilities, our method achieves higher actual win rates than baselines, even if it does not always select the strictly optimal engine.

In BrowseComp, we observe a similar pattern of improvement. Notably, for GPT-5 in the full search setting, \methodname recovers a significant portion of the performance gap (21.8\% $\rightarrow$ 22.1\%), whereas baselines like Play2Prompt and EasyTool actively degrade performance (15.8\% and 17.1\% respectively). We observe a single outlier in the Domain-specific (9) setting with the smaller model (GPT-5-mini), where EasyTool outperforms our method (7.1\% vs 3.2\%). However, this appears to be a stochastic anomaly, as EasyTool \textit{degrades} performance for the same model in the harder Full Search setting (9.7\% $\rightarrow$ 8.2\%), whereas \methodname significantly improves it (9.7\% $\rightarrow$ 12.8\%). This suggests that EasyTool's static summarization simply produced a prompt structure that coincidentally aided the smaller model in the simpler setting, but failed to generalize.

Overall, while \methodname establishes a robust baseline by demonstrating that reflecting on real trajectories drives better alignment than isolated processing, the substantial performance gap remaining compared to the Gold Oracle highlights the significant headroom available for future methods to further improve opaque tool reasoning.

\subsection{Analysis}

\begin{table}[t]
\footnotesize
\centering
\caption{Performance comparison table on BrowseComp Domains using Qwen-0.6B. Acc. refers to accuracy and \#TC refers to number of tool calls. \methodnameshort denotes \methodname, P2P denotes Play2Prompt,  and ET denotes EasyTool. The highest method Acc. value in each row is \textbf{bolded}.}
\vspace{-5pt}
\label{tab:browsecomp_full} 
\begin{tabular}{lcccccccccc}
\toprule
\multirow{3}{*}{\textbf{Model}} & \multicolumn{10}{c}{\textbf{ReAct}} \\
\cmidrule(l){2-11}
 & \multicolumn{2}{c}{\textbf{Gold}} & \multicolumn{2}{c}{\textbf{Base}} & \multicolumn{2}{c}{\textbf{+ \methodnameshort}} & \multicolumn{2}{c}{\textbf{+ P2P}} & \multicolumn{2}{c}{\textbf{+ ET}} \\
\cmidrule(lr){2-3} \cmidrule(lr){4-5} \cmidrule(lr){6-7} \cmidrule(lr){8-9} \cmidrule(l){10-11}
 & \textbf{Acc.} & \textbf{\#TC} & \textbf{Acc.} & \textbf{\#TC} & \textbf{Acc.} & \textbf{\#TC} & \textbf{Acc.} & \textbf{\#TC} & \textbf{Acc.} & \textbf{\#TC} \\
\midrule

\rowcolor{gray!20}
\multicolumn{11}{c}{Tool Setting: Domain-specific (9) Search} \\
GPT-5 & 22.5 & 7.0  & 18.1 & 10.0 & \textbf{21.9} & 6.3 & 17.9 & 7.8 & 18.9 & 8.2 \\
GPT-5-mini & 9.7 & 2.8 & 3.1 & 4.1 & 3.2 & 3.5  & 2.8 & 3.5 & \textbf{7.1} & 3.8\\
\midrule

\rowcolor{gray!20}
\multicolumn{11}{c}{Tool Setting: Domain-specific (9) + Full Search} \\
GPT-5 & 27.8 &  6.8 & 21.8 &  9.3 & \textbf{22.1} & 9.5 & 15.8 & 7.2 & 17.1 & 8.6 \\
GPT-5-mini & 12.6 &  3.0 & 9.7 & 3.7 & \textbf{12.8} & 3.3 & 5.6 & 3.5 & 8.2 & 3.5 \\
\bottomrule
\end{tabular}
\vspace{-5mm}
\end{table}
\begin{wrapfigure}{r}{0.55\textwidth} %
\centering
\vspace{-12pt}
\makebox[\linewidth][c]{%
\resizebox{1.02\linewidth}{!}{
\begin{tikzpicture}
\begin{axis}[
    name=plotA,  %
    xlabel={Iteration},
    ylabel={Acc},
    grid=major,
    title={A)},
        legend style={
        at={(1.7,1.7)},
        anchor=north,
        legend columns=3,
        /tikz/every even column/.append style={column sep=0.5cm}
    },
    width=0.3\textwidth,  %
    height=0.25\textwidth   %
]

\addplot[
    color=blue,
    mark=*,
    opacity=0.7,
    line width=1pt,
    mark size=2pt
] coordinates {
    (0, 0) (1,.34) (2,.54) (3,.54) (4,.60) (5,.62) (6,.62)
};
\addlegendentry{GPT-5-Mini}

\addplot[
    color=green!70!black,
    mark=*,
    opacity=0.7,
    line width=1pt,
    mark size=2pt
] coordinates {
    (0,0) (1,0.34) (2,0.56) (3,0.62) (4, 0.64) (5, 0.64) (6, 0.64) (7, 0.64) (8, 0.64) (9, 0.64)
};
\addlegendentry{GPT-5}

\end{axis}
\begin{axis}[
    name=plotB,
    at={($(plotA.east)+(0.75cm,0)$)},  %
    anchor=west,
    xlabel={Iteration},
    grid=major,
    title={B)},
        legend style={
        at={(0.5,-0.15)},
        anchor=north,
        legend columns=3,
        /tikz/every even column/.append style={column sep=0.5cm}
    },
    width=0.3\textwidth,  %
    height=0.25\textwidth   %
]

\addplot[
    color=blue,
    mark=*,
    opacity=0.7,
    line width=1pt,
    mark size=2pt
] coordinates {
    (0,0) (1,.82) (2,.82) (3,.82) (4,.82)
};

\addplot[
    color=green!70!black,
    mark=*,
    opacity=0.7,
    line width=1pt,
    mark size=2pt
] coordinates {
    (0,0) (1,.78) (2,.76) (3,.8) (4,.8) (5, .8)
};

\end{axis}
\begin{axis}[
    at={($(plotB.east)+(0.85cm,0)$)},  %
    anchor=west,
    xlabel={Iteration},
    grid=major,
    title={C)},
        legend style={
        at={(0.5,-0.15)},
        anchor=north,
        legend columns=3,
        /tikz/every even column/.append style={column sep=0.5cm}
    },
    width=0.3\textwidth,  %
    height=0.25\textwidth   %
]

\addplot[
    color=blue!70!black,
    mark=*,
    opacity=0.7,
    line width=1pt,
    mark size=2pt
] coordinates {
    (0,0.84) (1,0.86) (2,0.86) (3,0.86) (4,0.86)
};

\addplot[
    color=green!70!black,
    mark=*,
    opacity=0.7,
    line width=1pt,
    mark size=2pt
] coordinates {
    (0,0.74) (1,0.78) (2,0.78) (3,0.78) (4,0.78) (5,0.78) (6, 0.78) (7,0.8)
};

\end{axis}
\end{tikzpicture}
}}
\vspace{-15pt}
\caption{Performance of \methodname on BFCL-Opaque with increased iterations on the tool settings: A): Anon. Function Names, B): Anon. Function Names + Descriptions, C): Anon. Function Names + Param. Names. Iterations stop after full convergence. These are expanded versions of the \Cref{tab:bfcl_full} results.}
\label{fig:iterativeperformance}

\footnotesize
\centering
\captionof{table}{Average number of reflection iters. required by \methodname to converge on BFCL-Opaque.}
\label{tab:iterativeperformance}
\begin{tabular}{lcc}
    \toprule
    Configuration & GPT-5 & GPT-5-mini \\
    \midrule
    A) Anon. Fn. Names & 3.44 & 2.96 \\
    B) Anon. Fn. Names + Desc. & 2.66 & 2.60 \\
    C) Anon. Fn. Names + Param. & 2.22 & 2.24 \\
    \bottomrule
\end{tabular}
\vspace{-15pt}
\end{wrapfigure}

We analyze a few components of \methodname and our own \benchmarkname:

\textbf{\methodname performance over iterations}
\Cref{fig:iterativeperformance} shows that when minimal tool information is available (plot A), the model gradually improves across iterations, measured by execution accuracy on BFCL-Opaque. When some useful information is already provided (plot B), performance spikes early on then plateaus, with little additional gain. Finally, in plot C, where most of the important tool is available, improvements are negligible and performance saturates almost immediately. These patterns are consistent across all evaluated models. However, GPT-5 takes more iterations to converge at the final tool documentation, while GPT-5-Mini consistently converges sooner.

We also report the average reflection iterations required for convergence in \Cref{tab:iterativeperformance}. We observe two key trends. First, convergence speed correlates with information availability: as starting documentation improves (from anonymized names to including parameters), average iterations decrease by ${\approx}35\%$ (from $3.2$ to $2.2$). Second, we observe that in the most opaque setting, GPT-5 uses more iterations ($3.44$) than GPT-5-mini ($2.96$). Combined with the performance gap in \Cref{fig:iterativeperformance}, this suggests the stronger model engages in more thorough exploration to uncover complex behaviors. Even in the hardest setting, convergence occurs in under 3.5 iterations on average, demonstrating the experience-efficiency of \methodname.

\textbf{Fidelity of learned descriptions}
We assess the quality of generated documentation both quantitatively and qualitatively. First, we quantify the fidelity of learned documentation by measuring semantic (SBERT embedding \cite{Reimers2019SentenceBERTSE}) and lexical (ROUGE-1; \cite{Lin2004ROUGEAP}) similarity against gold standards on BFCL-Opaque (\Cref{tab:similarity_metrics}).  \methodname consistently outperforms baselines on both metrics. In the hardest ``Anon. Fn. Names'' setting, we achieve a semantic similarity of \textbf{0.58} (vs. 0.51 for Play2Prompt) and a lexical overlap of \textbf{0.28} (vs. 0.18). The improvement in both metrics confirms that \methodname captures both the general semantic meaning \textit{and} specific terminology and functional constraints from the gold documentation.

Second, we qualitatively analyze the descriptions generated for Chess tools in \Cref{appendix:FunctionDesc}. Interestingly, \methodname learns the nuances of \textit{when} each tool must be used. For example, the middle game specialist final description explicitly mentions the tool is useful for stabilizing dynamic middle games. A similar pattern is observed in the playing strength setting: the 1200 ELO specialist is described as effective at spotting immediate tactical conversions, while the 2400 ELO specialist is ideal for complex strategic maneuvering. On the other hand, Play2Prompt doesn't perform well in the strength differentiation setting -- with a shocking performance of 0.25 -- since the best tool's description (2400 ELO) remains under-specified. Instead, the agent always picks the \emph{second} best tool which has a well-specified tool description.
\begin{table}[t!]
  \centering
  \caption{Average similarity metrics (semantic and lexical) of final documentation generated vs. gold documentation on BFCL-Opaque. The highest sem. is bolded, while the highest lex. is italicized. \methodnameshort denotes \methodname, P2P denotes Play2Prompt, and ET denotes EasyTool.}
  \label{tab:similarity_metrics}
  \resizebox{\textwidth}{!}{%
    \begin{tabular}{lcccccccc}
    \toprule
     \multirow{2}{*}{\textbf{Documentation}}& \multicolumn{2}{c}{\textbf{\methodnameshort (GPT-5)}} & \multicolumn{2}{c}{\textbf{\methodnameshort (GPT-5-Mini)}} & \multicolumn{2}{c}{\textbf{P2P}} & \multicolumn{2}{c}{\textbf{ET}} \\
    \cmidrule(lr){2-3} \cmidrule(lr){4-5} \cmidrule(lr){6-7} \cmidrule(lr){8-9}
     & \textbf{Sem.} & \textbf{Lex.} & \textbf{Sem.} & \textbf{Lex.} & \textbf{Sem.} & \textbf{Lex.} & \textbf{Sem.} & \textbf{Lex.} \\
    \midrule

    Anon. Fn. Names & 0.44 & 0.21 & \textbf{0.58} & \textit{0.28} & 0.51 & 0.18 & 0   & 0  \\    
    
    Anon. Fn. Names + Desc.             & \textbf{0.78} & 0.43 & 0.78 & \textit{0.44} & 0.70 & 0.31 & 0.71 & 0.43 \\
    
    Anon. Fn. Names + Param. Names   & \textbf{0.71} & \textit{0.40} & 0.71 & \textit{0.40} & 0.69 & 0.28 & 0.69 & 0.39 \\

    \bottomrule
    \end{tabular}%
  }
  \vspace{-1mm}
\end{table} %
\begin{wraptable}{r}{0.5\textwidth}
\vspace{1mm}
\centering
\caption{Performance on BFCL-Opaque with different editors on Anon. Fn. Names setting for the executable multiple function subset}
\begin{tabular}{lccc}
\toprule
\textbf{LLM Agent} & \textbf{GPT5} & \textbf{GPT5-mini} & \textbf{o3} \\
\midrule
GPT5       & 0.64       & 0.62        & 0.60 \\
GPT5-mini  & 0.62       & 0.54        & 0.48 \\
\bottomrule
\vspace{-8mm}
\end{tabular}
\label{editorablation}
\end{wraptable} 
\textbf{Strength of LLM Agent vs Editor LLM}
We ablate the editor, using weaker models, i.e. GPT-5-Mini and O3, and measure the performance on BFCL in \Cref{editorablation}. The strength of the editor model directly affects the performance of both agents. Furthermore, the stronger agent model is generally more robust to a weak editor model. GPT-5-Mini sees a steep drop of 8 points when using GPT-mini as the editor. However, with GPT-5 as the agent and GPT-5-Mini editor, the observed drop is only 2 points.

\section{Related Work}

\textbf{LLM agents}
LLM Agents that can call tools enable interaction with external systems. \citet{Schick2023ToolformerLM} showed language models can learn to use tools through self-supervised learning, while ReAct \citep{Yao2022ReActSR} introduced the reasoning-action paradigm where agents alternate between thinking and acting. \citet{shinn2023reflexion} further enhanced tool-using agents with the ability to learn from failures through verbal self-reflection. These advances have enabled complex agent behaviors, from multi-agent coordination \citep{park2023generativeagentsinteractivesimulacra} to continuous skill acquisition \citep{wang2023voyageropenendedembodiedagent}. %

\textbf{Tool Documentation}
\citet{Hsieh2023ToolDE} demonstrate the critical role of comprehensive documentation in tool learning. Recent work has focused on automatically refining these descriptions: \citet{Qu2024FromET} introduces the DRAFT framework where gathered experience is used to rewrite tool documentation, while \citet{Fang2025PLAY2PROMPTZT} introduce Play2Prompt, which employs self-play followed by self-reflection to iteratively improve tool documentation. \citet{Yuan2024EASYTOOLEL} propose EASYTOOL, showing that condensing verbose descriptions and adding usage examples significantly improves downstream performance by reducing hallucination rates. \citet{Wang2024LLMsIT} extend this line of work by incorporating short-term and long-term memory mechanisms after self-reflection phases. %

\textbf{Tool Evaluation Benchmarks}
Current tool-use benchmarks include BFCL (Berkeley Function Calling Leaderboard), which evaluates simple, single-turn function calls, ToolBench \citep{xu2023toolmanipulationcapabilityopensource} and StableToolBench \citep{guo2024stabletoolbench}, which offer thousands of real-world REST APIs. However, ToolBench and StableToolBench suffer from API key accessibility issues and poor reproducibility due to their reliance on external services. %

\vspace{-1mm}
\section{Conclusion}
We introduce \benchmarkname, a benchmark consisting of three goal-oriented environments with opaque tools, which better reflects the reality of working with underspecified, poorly documented tools in real-world scenarios.
Through our proposed \methodname{} framework, we demonstrate that LLM agents can effectively learn to improve tool documentation through iterative observation of execution feedback, achieving superior performance while being significantly more token-efficient than existing approaches.

\bibliography{iclr2025_conference}

@article{Fang2025PLAY2PROMPTZT,
  title={PLAY2PROMPT: Zero-shot Tool Instruction Optimization for LLM Agents via Tool Play},
  author={Wei-Wen Fang and Yang Zhang and Kaizhi Qian and James Glass and Yada Zhu},
  journal={ArXiv},
  year={2025},
  volume={abs/2503.14432},
  url={https://api.semanticscholar.org/CorpusID:277104481}
}

@inproceedings{Radford2019LanguageMA,
  title={Language Models are Unsupervised Multitask Learners},
  author={Alec Radford and Jeff Wu and Rewon Child and David Luan and Dario Amodei and Ilya Sutskever},
  year={2019},
  url={https://api.semanticscholar.org/CorpusID:160025533}
}

@article{shinn2023reflexion,
  title={Reflexion: Language agents with verbal reinforcement learning},
  author={Shinn, Noah and Cassano, Federico and Gopinath, Ashwin and Narasimhan, Karthik and Yao, Shunyu},
  journal={Advances in Neural Information Processing Systems},
  volume={36},
  pages={8634--8652},
  year={2023}
}

@misc{guo2024stabletoolbench,
  title={StableToolBench: Towards Stable Large-Scale Benchmarking on Tool Learning of Large Language Models}, 
  author={Zhicheng Guo and Sijie Cheng and Hao Wang and Shihao Liang and Yujia Qin and Peng Li and Zhiyuan Liu and Maosong Sun and Yang Liu},
  year={2024},
  eprint={2403.07714},
  archivePrefix={arXiv},
  primaryClass={cs.CL}
}

@misc{xu2023toolmanipulationcapabilityopensource,
      title={On the Tool Manipulation Capability of Open-source Large Language Models}, 
      author={Qiantong Xu and Fenglu Hong and Bo Li and Changran Hu and Zhengyu Chen and Jian Zhang},
      year={2023},
      eprint={2305.16504},
      archivePrefix={arXiv},
      primaryClass={cs.CL},
      url={https://arxiv.org/abs/2305.16504}, 
}

@misc{wang2023voyageropenendedembodiedagent,
      title={Voyager: An Open-Ended Embodied Agent with Large Language Models}, 
      author={Guanzhi Wang and Yuqi Xie and Yunfan Jiang and Ajay Mandlekar and Chaowei Xiao and Yuke Zhu and Linxi Fan and Anima Anandkumar},
      year={2023},
      eprint={2305.16291},
      archivePrefix={arXiv},
      primaryClass={cs.AI},
      url={https://arxiv.org/abs/2305.16291}, 
}

@article{Wang2023ASO,
  title={A Survey on Large Language Model based Autonomous Agents},
  author={Lei Wang and Chengbang Ma and Xueyang Feng and Zeyu Zhang and Hao-ran Yang and Jingsen Zhang and Zhi-Yang Chen and Jiakai Tang and Xu Chen and Yankai Lin and Wayne Xin Zhao and Zhewei Wei and Ji-rong Wen},
  journal={ArXiv},
  year={2023},
  volume={abs/2308.11432},
  url={https://api.semanticscholar.org/CorpusID:261064713}
}

@misc{park2023generativeagentsinteractivesimulacra,
      title={Generative Agents: Interactive Simulacra of Human Behavior}, 
      author={Joon Sung Park and Joseph C. O'Brien and Carrie J. Cai and Meredith Ringel Morris and Percy Liang and Michael S. Bernstein},
      year={2023},
      eprint={2304.03442},
      archivePrefix={arXiv},
      primaryClass={cs.HC},
      url={https://arxiv.org/abs/2304.03442}, 
}

@article{Reimers2019SentenceBERTSE,
  title={Sentence-BERT: Sentence Embeddings using Siamese BERT-Networks},
  author={Nils Reimers and Iryna Gurevych},
  journal={ArXiv},
  year={2019},
  volume={abs/1908.10084},
  url={https://api.semanticscholar.org/CorpusID:201646309}
}

@inproceedings{Lin2004ROUGEAP,
  title={ROUGE: A Package for Automatic Evaluation of Summaries},
  author={Chin-Yew Lin},
  booktitle={Annual Meeting of the Association for Computational Linguistics},
  year={2004},
  url={https://api.semanticscholar.org/CorpusID:964287}
}

@article{Qu2024FromET,
  title={From Exploration to Mastery: Enabling LLMs to Master Tools via Self-Driven Interactions},
  author={Changle Qu and Sunhao Dai and Xiaochi Wei and Hengyi Cai and Shuaiqiang Wang and Dawei Yin and Jun Xu and Jirong Wen},
  journal={ArXiv},
  year={2024},
  volume={abs/2410.08197},
  url={https://api.semanticscholar.org/CorpusID:273233320}
}

@article{Schick2023ToolformerLM,
  title={Toolformer: Language Models Can Teach Themselves to Use Tools},
  author={Timo Schick and Jane Dwivedi-Yu and Roberto Dess{\`i} and Roberta Raileanu and Maria Lomeli and Luke Zettlemoyer and Nicola Cancedda and Thomas Scialom},
  journal={ArXiv},
  year={2023},
  volume={abs/2302.04761},
  url={https://api.semanticscholar.org/CorpusID:256697342}
}

@article{Shen2024ShortcutsBenchAL,
  title={ShortcutsBench: A Large-Scale Real-world Benchmark for API-based Agents},
  author={Haiyang Shen and Yue Li and Desong Meng and Dongqi Cai and Sheng Qi and Li Zhang and Mengwei Xu and Yun Ma},
  journal={ArXiv},
  year={2024},
  volume={abs/2407.00132},
  url={https://api.semanticscholar.org/CorpusID:270870543}
}

@article{Yao2022ReActSR,
  title={ReAct: Synergizing Reasoning and Acting in Language Models},
  author={Shunyu Yao and Jeffrey Zhao and Dian Yu and Nan Du and Izhak Shafran and Karthik Narasimhan and Yuan Cao},
  journal={ArXiv},
  year={2022},
  volume={abs/2210.03629},
  url={https://api.semanticscholar.org/CorpusID:252762395}
}

@software{stockfish,
  title = {Stockfish: Open Source Chess Engine},
  author = {Romstad, Tord and Costalba, Marco and Kiiski, Joona and others},
  year = {2024},
  url = {https://github.com/official-stockfish/Stockfish},
  version = {16}
}

@misc{openai2025introducinggpt5,
  author       = {OpenAI},
  title        = {Introducing GPT-5},
  year         = {2025},
  howpublished = {OpenAI},
  month        = aug,
  day          = {7},
  url          = {https://openai.com/index/introducing-gpt-5/}
}

@inproceedings{Zhang2025CompleteCG,
  title={Complete Chess Games Enable LLM Become A Chess Master},
  author={Yinqi Zhang and Xintian Han and Haolong Li and Kedi Chen and Shaohui Lin},
  booktitle={North American Chapter of the Association for Computational Linguistics},
  year={2025},
  url={https://api.semanticscholar.org/CorpusID:275954108}
}

@article{Qin2023ToolLLMFL,
  title={ToolLLM: Facilitating Large Language Models to Master 16000+ Real-world APIs},
  author={Yujia Qin and Shi Liang and Yining Ye and Kunlun Zhu and Lan Yan and Ya-Ting Lu and Yankai Lin and Xin Cong and Xiangru Tang and Bill Qian and Sihan Zhao and Runchu Tian and Ruobing Xie and Jie Zhou and Marc H. Gerstein and Dahai Li and Zhiyuan Liu and Maosong Sun},
  journal={ArXiv},
  year={2023},
  volume={abs/2307.16789},
  url={https://api.semanticscholar.org/CorpusID:260334759}
}

@article{Mialon2023AugmentedLM,
  title={Augmented Language Models: a Survey},
  author={Gr{\'e}goire Mialon and Roberto Dess{\`i} and Maria Lomeli and Christoforos Nalmpantis and Ramakanth Pasunuru and Roberta Raileanu and Baptiste Rozi{\`e}re and Timo Schick and Jane Dwivedi-Yu and Asli Celikyilmaz and Edouard Grave and Yann LeCun and Thomas Scialom},
  journal={Trans. Mach. Learn. Res.},
  year={2023},
  volume={2023},
  url={https://api.semanticscholar.org/CorpusID:256868474}
}

@inproceedings{guo-etal-2024-stabletoolbench,
    title = "{S}table{T}ool{B}ench: Towards Stable Large-Scale Benchmarking on Tool Learning of Large Language Models",
    author = "Guo, Zhicheng  and
      Cheng, Sijie  and
      Wang, Hao  and
      Liang, Shihao  and
      Qin, Yujia  and
      Li, Peng  and
      Liu, Zhiyuan  and
      Sun, Maosong  and
      Liu, Yang",
    editor = "Ku, Lun-Wei  and
      Martins, Andre  and
      Srikumar, Vivek",
    booktitle = "Findings of the Association for Computational Linguistics: ACL 2024",
    month = aug,
    year = "2024",
    address = "Bangkok, Thailand",
    publisher = "Association for Computational Linguistics",
    url = "https://aclanthology.org/2024.findings-acl.664/",
    doi = "10.18653/v1/2024.findings-acl.664",
    pages = "11143--11156"
}

@article{Patil2023GorillaLL,
  title={Gorilla: Large Language Model Connected with Massive APIs},
  author={Shishir G. Patil and Tianjun Zhang and Xin Wang and Joseph E. Gonzalez},
  journal={ArXiv},
  year={2023},
  volume={abs/2305.15334},
  url={https://api.semanticscholar.org/CorpusID:258865184}
}

@article{Brown2020LanguageMA,
  title={Language Models are Few-Shot Learners},
  author={Tom B. Brown and Benjamin Mann and Nick Ryder and Melanie Subbiah and Jared Kaplan and Prafulla Dhariwal and Arvind Neelakantan and Pranav Shyam and Girish Sastry and Amanda Askell and Sandhini Agarwal and Ariel Herbert-Voss and Gretchen Krueger and T. J. Henighan and Rewon Child and Aditya Ramesh and Daniel M. Ziegler and Jeff Wu and Clemens Winter and Christopher Hesse and Mark Chen and Eric Sigler and Ma-teusz Litwin and Scott Gray and Benjamin Chess and Jack Clark and Christopher Berner and Sam McCandlish and Alec Radford and Ilya Sutskever and Dario Amodei},
  journal={ArXiv},
  year={2020},
  volume={abs/2005.14165},
  url={https://api.semanticscholar.org/CorpusID:218971783}
}

@article{Yuan2024EASYTOOLEL,
  title={EASYTOOL: Enhancing LLM-based Agents with Concise Tool Instruction},
  author={Siyu Yuan and Kaitao Song and Jiangjie Chen and Xu Tan and Yongliang Shen and Ren Kan and Dongsheng Li and Deqing Yang},
  journal={ArXiv},
  year={2024},
  volume={abs/2401.06201},
  url={https://api.semanticscholar.org/CorpusID:266977201}
}

@article{Hsieh2023ToolDE,
  title={Tool Documentation Enables Zero-Shot Tool-Usage with Large Language Models},
  author={Cheng-Yu Hsieh and Sibei Chen and Chun-Liang Li and Yasuhisa Fujii and Alexander J. Ratner and Chen-Yu Lee and Ranjay Krishna and Tomas Pfister},
  journal={ArXiv},
  year={2023},
  volume={abs/2308.00675},
  url={https://api.semanticscholar.org/CorpusID:260351459}
}

@inproceedings{Wang2024LLMsIT,
  title={LLMs in the Imaginarium: Tool Learning through Simulated Trial and Error},
  author={Boshi Wang and Hao Fang and Jason Eisner and Benjamin Van Durme and Yu Su},
  booktitle={Annual Meeting of the Association for Computational Linguistics},
  year={2024},
  url={https://api.semanticscholar.org/CorpusID:268264353}
}

@inproceedings{Chen2025BrowseCompPlusAM,
  title={BrowseComp-Plus: A More Fair and Transparent Evaluation Benchmark of Deep-Research Agent},
  author={Zijian Chen and Xueguang Ma and Shengyao Zhuang and Ping Nie and Kai Zou and Andrew Liu and Joshua Green and Kshama Patel and Ruoxi Meng and Mingyi Su and Sahel Sharifymoghaddam and Yanxi Li and Haoran Hong and Xinyu Shi and Xuye Liu and Nandan Thakur and Crystina Zhang and Luyu Gao and Wenhu Chen and Jimmy Lin},
  year={2025},
  url={https://api.semanticscholar.org/CorpusID:280565737}
}

@inproceedings{patil2025bfcl,
title={The Berkeley Function Calling Leaderboard (BFCL): From Tool Use to Agentic Evaluation of Large Language Models}, 
author={Patil, Shishir G. and Mao, Huanzhi and Cheng-Jie Ji, Charlie and Yan, Fanjia and Suresh, Vishnu and Stoica, Ion and E. Gonzalez, Joseph},
booktitle={Forty-second International Conference on Machine Learning},
year={2025},
}
\bibliographystyle{iclr2025_conference}

\newpage

\appendix
\section{Experimental Details}
\label{appendix:experiments}

We list further experimental details and results in this section.

\subsection{\methodname Hyperparameters}

For all datasets, we use the ReAct \citep{Yao2022ReActSR} framework, which prompts the agent to generate interleaved reasoning traces and executable tool calls to solve tasks dynamically. 

\textbf{Max iterations}: For BFCL Opaque, we use a max iterations of 10 to iteratively test and optimize tool descriptions. For BrowseComp and Chess, we use a max iterations of 3 due to computational constraints (these trajectories are much larger and we have many more of them). Recall that for BFCL, the LLM agent can stop early if it reaches a final answer; we report the average number of iterations taken in \Cref{tab:bfcl_full}.

\textbf{Offline Batch Size}  For the two tasks which we use the offline version of \methodname for - Chess and BrowseComp Domains -- we use a mini-batch size of 10 trajectories.

\textbf{Chess Specifics:} 
To ensure computational efficiency during the \emph{exploration phase}, we simulate partial trajectories of 10 moves starting from the sampled position. We evaluate the state at the end of this partial trajectory using the board value estimate from Stockfish \citep{stockfish}. This scalar value acts as the dense process reward described in \S 4.1, allowing the editor language model to critique the strategic quality of tool choices without needing to reach a checkmate. 

\textbf{Generation Details}
Our experiments are primarily done with GPT-5 and GPT-5-mini \footnote{\url{https://platform.openai.com/docs/models/gpt-5} and \url{https://platform.openai.com/docs/models/gpt-5-mini}}; we sample from these models (OpenAI does not allow setting temperature or top-p parameters for these models). To reduce latency during exploration on Chess which has long trajectories, we set reasoning effort to be minimal. Everywhere else, we use "medium" reasoning effort, including on the documentation updates.

Finally, in Figures \ref{fig:chess_exploration} - \ref{fig:bc_merge} in \Cref{appendix:prompts}, we detail the prompts we use across our three datasets for  1) tool usage/exploration 2) documentation updates and 3) the prompts for merging the learned descriptions (if applicable).

\subsection{\benchmarkname Dataset Details}

\textbf{BrowseComp Domains}
We run BrowseComp Domains using the Qwen-0.6B embedder model, following 
\citealp{Chen2025BrowseCompPlusAM}. Also following this work, we set the max number of tokens in retrieved snippets to be 512, and the number of retrieved documents per search tool to be $k=5$.

\textbf{Chess}
For the \emph{final evaluation} on the test set (\Cref{tab:chess_full}), we simulate full games. We enforce a maximum limit of \textbf{120 moves}; if the game exceeds this limit without a decisive result (checkmate/stalemate), it is recorded as a draw.

\section{Comparison of Function Descriptions For Chess Tools}
\label{appendix:FunctionDesc}

\subsection{Ground Truth}
The following are the ground truth descriptions generated for different chess tools.

\begin{tcolorbox}[
    colback=gray!5!white,
    colframe=gray!75!black,
    title=\textbf{Opening Specialist},
    boxrule=1pt,
    left=3mm,
    right=3mm,
    top=2mm,
    bottom=2mm,
    sharp corners,
    fonttitle=\bfseries\color{white},
    colbacktitle=gray!75!black,
    width=0.95\textwidth
]
Analyzes the current board position and returns a suggested move in algebraic notation. Returns a 2400 Elo strength move if the current position is in the opening phase (26+ pieces remaining, excluding kings), otherwise returns a random legal move.
\end{tcolorbox}

\vspace{0.3em}

\begin{tcolorbox}[
    colback=gray!5!white,
    colframe=gray!75!black,
    title=\textbf{Middle Game Specialist},
    boxrule=1pt,
    left=3mm,
    right=3mm,
    top=2mm,
    bottom=2mm,
    sharp corners,
    fonttitle=\bfseries\color{white},
    colbacktitle=gray!75!black,
    width=0.95\textwidth
]
Analyzes the current board position and returns a suggested move in algebraic notation. Returns a 2400 Elo strength move if the current position is in the middlegame phase (14-25 pieces remaining, excluding kings), otherwise returns a random legal move.
\end{tcolorbox}

\vspace{0.3em}

\begin{tcolorbox}[
    colback=gray!5!white,
    colframe=gray!75!black,
    title=\textbf{End Game Specialist},
    boxrule=1pt,
    left=3mm,
    right=3mm,
    top=2mm,
    bottom=2mm,
    sharp corners,
    fonttitle=\bfseries\color{white},
    colbacktitle=gray!75!black,
    width=0.95\textwidth
]
Analyzes the current board position and returns a suggested move in algebraic notation. Returns a 2400 Elo strength move if the current position is in the endgame phase (8-13 pieces remaining, excluding kings), otherwise returns a random legal move.
\end{tcolorbox}

\vspace{0.3em}

\begin{tcolorbox}[
    colback=gray!5!white,
    colframe=gray!75!black,
    title=\textbf{Late Game Specialist},
    boxrule=1pt,
    left=3mm,
    right=3mm,
    top=2mm,
    bottom=2mm,
    sharp corners,
    fonttitle=\bfseries\color{white},
    colbacktitle=gray!75!black,
    width=0.95\textwidth
]
Analyzes the current board position and returns a suggested move in algebraic notation. Returns a 2400 Elo strength move if the current position is in the late endgame phase (fewer than 8 pieces remaining, excluding kings), otherwise returns a random legal move.
\end{tcolorbox}

\vspace{0.3em}

\begin{tcolorbox}[
    colback=gray!5!white,
    colframe=gray!75!black,
    title=\textbf{1200 ELO Specialist},
    boxrule=1pt,
    left=3mm,
    right=3mm,
    top=2mm,
    bottom=2mm,
    sharp corners,
    fonttitle=\bfseries\color{white},
    colbacktitle=gray!75!black,
    width=0.95\textwidth
]
Analyzes the current board position and returns a suggested move in algebraic notation. The tool automatically reads the board state — no input is needed. Suggested move is calculated at 1200 Elo strength.
\end{tcolorbox}

\vspace{0.3em}

\begin{tcolorbox}[
    colback=gray!5!white,
    colframe=gray!75!black,
    title=\textbf{1800 ELO Specialist},
    boxrule=1pt,
    left=3mm,
    right=3mm,
    top=2mm,
    bottom=2mm,
    sharp corners,
    fonttitle=\bfseries\color{white},
    colbacktitle=gray!75!black,
    width=0.95\textwidth
]
Analyzes the current board position and returns a suggested move in algebraic notation. The tool automatically reads the board state — no input is needed. Suggested move is calculated at 1800 Elo strength.
\end{tcolorbox}

\vspace{0.3em}

\begin{tcolorbox}[
    colback=gray!5!white,
    colframe=gray!75!black,
    title=\textbf{2400 ELO Specialist},
    boxrule=1pt,
    left=3mm,
    right=3mm,
    top=2mm,
    bottom=2mm,
    sharp corners,
    fonttitle=\bfseries\color{white},
    colbacktitle=gray!75!black,
    width=0.95\textwidth
]
Analyzes the current board position and returns a suggested move in algebraic notation. The tool automatically reads the board state — no input is needed. Suggested move is calculated at 2400 Elo strength.
\end{tcolorbox}

\vspace{0.3em}

\subsection{\methodname}
The following are the descriptions generated by our method for different chess tools.

\begin{tcolorbox}[
    colback=blue!5!white,
    colframe=blue!75!black,
    title=\textbf{Opening Specialist},
    boxrule=1pt,
    left=3mm,
    right=3mm,
    top=2mm,
    bottom=2mm,
    sharp corners,
    fonttitle=\bfseries\color{white},
    colbacktitle=blue!75!black,
    width=0.95\textwidth
]
Best as a consolidation tool when already worse or in simplified positions, especially endgames. It favors quiet, low-variance moves that limit further damage. Less suitable for seizing the initiative or equalizing from balanced positions, and it can misjudge in sharp middlegames or make superficial material grabs in quiet endings. Use it to steady the ship, not to press for precise gains.
\end{tcolorbox}

\vspace{0.3em}

\begin{tcolorbox}[
    colback=blue!5!white,
    colframe=blue!75!black,
    title=\textbf{Middle Game Specialist},
    boxrule=1pt,
    left=3mm,
    right=3mm,
    top=2mm,
    bottom=2mm,
    sharp corners,
    fonttitle=\bfseries\color{white},
    colbacktitle=blue!75!black,
    width=0.95\textwidth
]
Best for stabilizing dynamic middlegames and defusing pressure through sound structural choices, particularly when defending. In calm or technical endgames and balanced positions, it tends toward passivity or inaccurate king/pawn decisions and is not a precise converter. Choose it to neutralize threats, not to fine-tune technical outcomes.
\end{tcolorbox}

\vspace{0.3em}

\begin{tcolorbox}[
    colback=blue!5!white,
    colframe=blue!75!black,
    title=\textbf{End Game Specialist},
    boxrule=1pt,
    left=3mm,
    right=3mm,
    top=2mm,
    bottom=2mm,
    sharp corners,
    fonttitle=\bfseries\color{white},
    colbacktitle=blue!75!black,
    width=0.95\textwidth
]
Best in simplified, technical positions that reward precision -- especially rook endgames -- and when an accurate, active move is needed to maintain control from an equal or slightly worse stance. Avoid it in sharp, tactical middlegames or when low-risk, resilient defense is required; it can overreach and swing the evaluation heavily. It's also not ideal for clinging on in lost endgames as White.
\end{tcolorbox}

\vspace{0.3em}

\begin{tcolorbox}[
    colback=blue!5!white,
    colframe=blue!75!black,
    title=\textbf{Late Game Specialist},
    boxrule=1pt,
    left=3mm,
    right=3mm,
    top=2mm,
    bottom=2mm,
    sharp corners,
    fonttitle=\bfseries\color{white},
    colbacktitle=blue!75!black,
    width=0.95\textwidth
]
Best for holding calm, technical endgames and maximizing resistance with safe, non-committal moves. Serviceable as a middle option when the position isn't tactically charged. Avoid it in sharp or tactical situations and whenever king safety is delicate; it tends to overlook immediate threats and can trigger large single-step collapses.
\end{tcolorbox}

\vspace{0.3em}

\begin{tcolorbox}[
    colback=blue!5!white,
    colframe=blue!75!black,
    title=\textbf{1200 ELO Specialist},
    boxrule=1pt,
    left=3mm,
    right=3mm,
    top=2mm,
    bottom=2mm,
    sharp corners,
    fonttitle=\bfseries\color{white},
    colbacktitle=blue!75!black,
    width=0.95\textwidth
]
Best when you want a decisive, forcing solution. Excels at spotting immediate conversions -- tactical captures, central pawn breaks, direct mating or queening threats -- and at jump-starting counterplay through rapid coordination (e.g., quick castling, active piece placement). Ideal for positions where concrete calculation can resolve tension right away. Tends to be high-variance: strong at seizing chances, but can overpress or misjudge tactical safety in messy defenses. Less suited to slow consolidation or risk-minimizing play.
\end{tcolorbox}

\vspace{0.3em}

\begin{tcolorbox}[
    colback=blue!5!white,
    colframe=blue!75!black,
    title=\textbf{1800 ELO Specialist},
    boxrule=1pt,
    left=3mm,
    right=3mm,
    top=2mm,
    bottom=2mm,
    sharp corners,
    fonttitle=\bfseries\color{white},
    colbacktitle=blue!75!black,
    width=0.95\textwidth
]
Best for building and sustaining initiative with forcing play. Excels at creating and maintaining pressure through checks, rook lifts, and energetic pawn breaks, and will alter the structure when it strengthens activity. Strong at converting an edge by keeping the opponent on the back foot. Less reliable when the position demands quiet consolidation or a concrete defensive neutralization; can overlook the need to stabilize before pressing.
\end{tcolorbox}

\vspace{0.3em}

\begin{tcolorbox}[
    colback=blue!5!white,
    colframe=blue!75!black,
    title=\textbf{2400 ELO Specialist},
    boxrule=1pt,
    left=3mm,
    right=3mm,
    top=2mm,
    bottom=2mm,
    sharp corners,
    fonttitle=\bfseries\color{white},
    colbacktitle=blue!75!black,
    width=0.95\textwidth
]
Best for stabilization and neutralization. Excels at consolidating king safety, coordinating pieces, recapturing accurately, and simplifying when under pressure. Finds direct defensive resources -- exchanges and file contests -- that reduce the opponent's practical chances. Ideal in worse or volatile positions and whenever risk control is paramount. Tends to forgo sharp attacking continuations in favor of solid, resilient play.
\end{tcolorbox}

\subsection{Play2Prompt}
The following are the descriptions generated by Play2Prompt for different chess tools.

\begin{tcolorbox}[
    colback=red!5!white,
    colframe=red!75!black,
    title=\textbf{Opening Specialist},
    boxrule=1pt,
    left=3mm,
    right=3mm,
    top=2mm,
    bottom=2mm,
    sharp corners,
    fonttitle=\bfseries\color{white},
    colbacktitle=red!75!black,
    width=1\textwidth
]
function\_1 analyzes a chess position from a FEN string and returns a single best-move suggestion in SAN (Standard Algebraic Notation).

\vspace{0.2em}

\textbf{Parameters}

\vspace{0.2em}

\begin{itemize}[leftmargin=*]
  \item \texttt{board\_state} (string, required): Complete 6-field FEN: \texttt{<pieces> <active> <castling> <en-passant> <halfmove> <fullmove>}. Provide a legal position and include all fields, separated by single spaces.  
  Example: \texttt{rnbqkbnr/pppppppp/8/8/8/8/PPPPPPPP/RNBQKBNR w KQkq - 0 1}
\end{itemize}

\vspace{0.2em}

\textbf{Output}

\vspace{0.2em}

\begin{itemize}[leftmargin=*]
  \item A SAN move string, e.g., ``e4'', ``Nf3'', ``exd5'', ``O-O'', ``O-O-O'', ``Bb5+'', ``Qh7\#''.
\end{itemize}

\vspace{0.2em}

\textbf{Usage example}

\vspace{0.2em}

\begin{itemize}[leftmargin=*]
  \item Call name: \texttt{function\_1} (do not use the original name).
  \item Input JSON: 
  
  \texttt{\{ "board\_state": "rnbqkbnr/pppppppp/8/8/8/8/PPPPPPPP/RNBQKBNR w KQkq - 0 1" \}}
  \item Example result: ``e4''
\end{itemize}

\vspace{0.2em}

\textbf{Notes}

\vspace{0.2em}

\begin{itemize}
  \item Use the exact key \texttt{board\_state}.
  \item Output is SAN only (not UCI/LAN).
\end{itemize}
\end{tcolorbox}

\vspace{0.3em}

\begin{tcolorbox}[
    colback=red!5!white,
    colframe=red!75!black,
    title=\textbf{Middle Game Specialist},
    boxrule=1pt,
    left=3mm,
    right=3mm,
    top=2mm,
    bottom=2mm,
    sharp corners,
    fonttitle=\bfseries\color{white},
    colbacktitle=red!75!black,
    width=1\textwidth
]
Analyzes a standard chess position and returns a single best move suggestion.

\vspace{0.2em}

\textbf{Input:}

A complete 6-field FEN string.

\vspace{0.2em}

\textbf{Parameter:}

\vspace{0.2em}

\begin{itemize}[leftmargin=*]
  \item \texttt{board\_state} (string, required) — Valid FEN with:
  \begin{enumerate}
    \item piece placement (8 ranks separated by ``/'', digits for empty squares, pieces PNBRQK/pnbrqk),
    \item active color ``w'' or ``b'',
    \item castling rights as a subset of ``KQkq'' or ``-'',
    \item en-passant target square ``-'' or a3–h6,
    \item halfmove clock (non-negative integer),
    \item fullmove number ($\geq$1).
  \end{enumerate}
\end{itemize}

\vspace{0.2em}

\textbf{Output:}

One move in Standard Algebraic Notation (SAN), e.g., ``e4'', ``Nf3'', ``exd5'', ``Qh5+'', ``e8=Q\#'', including disambiguation as needed. No extra text.

\vspace{0.2em}

\textbf{Scope:}

Standard chess only (not Chess960).

\vspace{0.2em}

\textbf{Validation:}

Malformed/illegal FEN may be rejected — ensure correct field count and values.

\vspace{0.2em}

\textbf{Examples:}

\begin{itemize}[leftmargin=*]
  \item \texttt{board\_state: "rnbqkbnr/pppppppp/8/8/4P3/8/PPPP1PPP/RNBQKBNR b KQkq - 0 1"} $\rightarrow$ ``d5''
  \item \texttt{board\_state: "r1bqkbnr/pppppppp/2n5/8/8/2N5/PPPPPPPP/R1BQKBNR w KQkq - 0 3"} $\rightarrow$ ``Nf3''
\end{itemize}

\end{tcolorbox}

\vspace{0.3em}

\begin{tcolorbox}[
    colback=red!5!white,
    colframe=red!75!black,
    title=\textbf{End Game Specialist},
    boxrule=1pt,
    left=3mm,
    right=3mm,
    top=2mm,
    bottom=2mm,
    sharp corners,
    fonttitle=\bfseries\color{white},
    colbacktitle=red!75!black,
    width=0.95\textwidth
]
Analyzes a chess position and returns a single move suggestion in Standard Algebraic Notation (SAN).  
Alias: \texttt{endgame\_specialist}.  
Input must be a complete, valid FEN string (six space-separated fields): piece placement, side to move (w/b), castling rights, en passant target square, halfmove clock, fullmove number.  
The side to move is taken from the FEN.

\vspace{0.2em}

\textbf{Parameters:}

\vspace{0.2em}

\begin{itemize}[leftmargin=*]
  \item \texttt{board\_state} (string, required): Full FEN for the current position. Include correct castling rights and en passant target if applicable.
\end{itemize}

\vspace{0.2em}

\textbf{Output:}

\vspace{0.2em}

\begin{itemize}[leftmargin=*]
  \item One SAN move string (not UCI/coordinate), e.g., ``Nf3'', ``exd5'', ``O-O'', ``O-O-O'', ``a8=Q'', ``Qh7+'', ``Rxf8\#'', with standard disambiguation as needed. Captures use ``x''; promotions use ``=Q/R/B/N''; checks ``+''; checkmates ``\#''.
\end{itemize}

\vspace{0.2em}

\textbf{Example:}

\begin{itemize}[leftmargin=*]
  \item \texttt{board\_state: "rnbqkbnr/pppppppp/8/8/8/8/PPPPPPPP/RNBQKBNR w KQkq - 0 1"} $\rightarrow$ ``e4''
\end{itemize}

\end{tcolorbox}

\vspace{0.3em}

\begin{tcolorbox}[
    colback=red!5!white,
    colframe=red!75!black,
    title=\textbf{Late Game Specialist},
    boxrule=1pt,
    left=3mm,
    right=3mm,
    top=2mm,
    bottom=2mm,
    sharp corners,
    fonttitle=\bfseries\color{white},
    colbacktitle=red!75!black,
    width=1.15\textwidth
]
Analyzes a chess position and returns a single move suggestion in Standard Algebraic Notation (SAN) as a plain string.

\vspace{0.2em}

\textbf{Parameters}

\vspace{0.2em}

\begin{itemize}[leftmargin=*]
  \item \texttt{board\_state} (string, required): Full 6-field FEN of the current position.  
  Format: ``<piece placement> <side to move> <castling rights> <en passant target> <halfmove clock> <fullmove number>''.  
  Example: \texttt{"rnbqkbnr/pppppppp/8/8/8/8/PPPPPPPP/RNBQKBNR w KQkq - 0 1"}.  
  Must represent a legal position.
\end{itemize}

\vspace{0.2em}

\textbf{Output}

\vspace{0.2em}

\begin{itemize}[leftmargin=*]
  \item SAN move string only (no JSON object).  
  Examples: ``e4'', ``Nf3'', ``exd5'', ``O-O'', ``O-O-O'', ``e8=Q'', ``Qh7\#'', ``Rd1+''.
\end{itemize}

\vspace{0.2em}

\textbf{Notes}

\vspace{0.2em}

\begin{itemize}[leftmargin=*]
  \item Input must be FEN (not PGN or UCI).  
  \item SAN uses uppercase piece letters (pawn omitted), ``x'' for captures, ``+/\#'' for check/mate, and ``=Q/R/B/N'' for promotions.
\end{itemize}

\vspace{0.2em}

\textbf{Example call}

\begin{itemize}[leftmargin=*]
  \item \texttt{function\_4(\{ "board\_state": "rnbqkbnr/pppppppp/8/8/8/8/PPPPPPPP/RNBQKBNR w KQkq - 0 1" \})}
\end{itemize}

\vspace{0.2em}

\textbf{Example response}

\begin{itemize}[leftmargin=*]
  \item ``e4''
\end{itemize}

\end{tcolorbox}

\vspace{0.3em}

\begin{tcolorbox}[
    colback=red!5!white,
    colframe=red!75!black,
    title=\textbf{ELO 1200 Specialist},
    boxrule=1pt,
    left=3mm,
    right=3mm,
    top=2mm,
    bottom=2mm,
    sharp corners,
    fonttitle=\bfseries\color{white},
    colbacktitle=red!75!black,
    width=0.95\textwidth
]
Suggests a single chess move for the side to move in the given position using a shallow fixed-depth analysis (\textasciitilde2 plies). Always call this tool when a move is requested from a FEN; do not infer moves without it.

\vspace{0.2em}

\textbf{Parameters:}

\vspace{0.2em}

\begin{itemize}[leftmargin=*]
  \item \texttt{board\_state} (string, required): A valid full FEN for standard chess with all 6 fields: piece placement, active color (w/b), castling rights (KQkq or -), en passant target (square or -), halfmove clock, fullmove number. The FEN must be legal and consistent; castling and en passant fields affect legality.
\end{itemize}

\vspace{0.2em}

\textbf{Output:}

A single move in Standard Algebraic Notation (SAN), not UCI/LAN. Examples: ``e4'', ``Nf3'', ``O-O'', ``R1e2'', ``exd5'', ``e8=Q\#''. Includes ``+'' or ``\#'' if applicable. No extra text.

\vspace{0.2em}

\textbf{Notes:}

If multiple moves are near-equal, one is returned.

\vspace{0.2em}

\textbf{Example:}

\texttt{board\_state="rnbqkbnr/pppppppp/8/8/8/8/PPPPPPPP/RNBQKBNR w KQkq - 0 1"} $\rightarrow$ ``e4''.

\end{tcolorbox}

\vspace{0.3em}

\begin{tcolorbox}[
    colback=red!5!white,
    colframe=red!75!black,
    title=\textbf{ELO 1800 Specialist},
    boxrule=1pt,
    left=3mm,
    right=3mm,
    top=2mm,
    bottom=2mm,
    sharp corners,
    fonttitle=\bfseries\color{white},
    colbacktitle=red!75!black,
    width=0.95\textwidth
]
Analyzes a standard chess position and returns the single best move at fixed search depth 8 plies. Use exactly one required parameter.

\vspace{0.2em}

\textbf{Required parameter:}

\vspace{0.2em}

\begin{itemize}[leftmargin=*]
  \item \texttt{board\_state} (string): A single-line valid FEN (Forsyth–Edwards Notation). Must include exactly 6 space-separated fields: 
  \begin{enumerate}
    \item piece placement,
    \item side to move (w/b),
    \item castling rights (KQkq or -),
    \item en passant target square (e.g., e3 or -),
    \item halfmove clock (integer),
    \item fullmove number (integer).
  \end{enumerate}
  No extra whitespace or newlines; standard chess only; position should be legal.
\end{itemize}

\vspace{0.2em}

\textbf{Returns:}

One move in SAN (Standard Algebraic Notation), e.g., ``Nf3'', ``exd5'', ``O-O'', ``O-O-O'', ``e8=Q'', ``Rxd8+'', ``Qh7\#'' (with disambiguation if needed). Not UCI; no scores or move lists.

\vspace{0.2em}

\textbf{Example call:}

\texttt{\{"board\_state":"rnbqkbnr/pppppppp/8/8/4P3/8/PPPP1PPP/RNBQKBNR b KQkq - 0 1"\}}

\vspace{0.2em}

\textbf{Example output format:}

``Nf6''

\end{tcolorbox}

\vspace{0.3em}

\begin{tcolorbox}[
    colback=red!5!white,
    colframe=red!75!black,
    title=\textbf{ELO 2400 Specialist},
    boxrule=1pt,
    left=3mm,
    right=3mm,
    top=2mm,
    bottom=2mm,
    sharp corners,
    fonttitle=\bfseries\color{white},
    colbacktitle=red!75!black,
    width=0.95\textwidth
]
Analyzes a standard chess position and returns the engine’s best move as a single SAN string.

\vspace{0.2em}

\textbf{Parameter:}

\vspace{0.2em}

\begin{itemize}[leftmargin=*]
  \item \texttt{board\_state} (string, required) — a Forsyth–Edwards Notation (FEN) of a legal position; include side-to-move and other fields (full 6-field FEN preferred). The move is determined from the FEN’s side-to-move.
\end{itemize}

\vspace{0.2em}

\textbf{Output:}

Exactly one move in Standard Algebraic Notation (SAN), e.g., ``e4'', ``Nf3'', ``O-O'', ``exd5'', ``Qh8\#'', ``b8=Q+'', ``R1e2''. Do not expect UCI/LAN or any extra text.

\vspace{0.2em}

\textbf{Notes:}

Fixed internal search depth (cannot be configured); standard chess only (no variants). Use the exact key name \texttt{board\_state} and pass only the FEN string.

\vspace{0.2em}

\textbf{Example call:}

\texttt{\{"board\_state":"rnbqkbnr/pppppppp/8/8/4P3/8/PPPP1PPP/RNBQKBNR b KQkq - 0 1"\}}

\vspace{0.2em}

\textbf{Example response:}

``d5''
\end{tcolorbox}

\vspace{0.3em}

\subsection{EasyTool}
The following is the description generated by EasyTool for chess tools; note that they are almost all identical since the initial description is the same (differences are only up to sample differences).

\begin{tcolorbox}[
    colback=green!5!white,
    colframe=green!75!black,
    title=\textbf{All tools},
    boxrule=1pt,
    left=3mm,
    right=3mm,
    top=2mm,
    bottom=2mm,
    sharp corners,
    fonttitle=\bfseries\color{white},
    colbacktitle=green!75!black,
    width=1.1\textwidth
]
\textbf{Tool Description:} \lq function\_1' analyzes a chess position from a FEN  \lq board\_state' and returns a suggested move in algebraic notation.

\vspace{0.3em}

\textbf{Tool Guidelines:}
\vspace{0.1em}

\textit{Scenario:}
\vspace{0.1em}

if you want to analyze a chess position (e.g., the Ruy Lopez after 4...Nf6) and get a suggested move for White.

\vspace{0.1em}
\textit{Parameters:} 

\vspace{0.1em}

\texttt{\{"board\_state":"r1bqkb1r/1ppp1ppp/p1n2n2/4p3/B3P3/5N2/PPPP1PPP/RNBQK2R w KQkq - 2 5"\}}

\end{tcolorbox}

\pagebreak

\section{Tasks}
\ashwin{IMPORTANT: The appendix has old stuff, need to do a pass to make it consistent with rest of the paper}
\subsection{BFCL-Opaque: Discovering Tool Functionality from Opaque Descriptions}
\label{appendix:BFCL}

The Berkeley Function Calling Leaderboard (BFCL) \citep{patil2025bfcl} \jack{give a more concrete description, like: "consists of 1K simple tasks like X; each task has between 1-3 functions like Y..."} %
We systematically degrade BFCL's tool descriptions to create an opaque setting where models must infer functionality through interaction. This tests the fundamental ability to map between ambiguous tool interfaces and their underlying behaviors. %

\paragraph{Task Setup:} Models receive user queries requiring specific tool calls (e.g., ``What's the weather in San Francisco?'', ``Schedule a meeting for tomorrow at 3pm'') but must discover which tools accomplish each task. We provide tools with systematically degraded documentation: function names are replaced with generic identifiers (e.g., \texttt{tool\_1}, \texttt{tool\_2}), descriptions are removed or made ambiguous, and parameter specifications lack type information or semantic hints. Models must experiment with different tools and parameter combinations to discover correct usage patterns.

\paragraph{Tool Degradation Strategy:} %
Initially, the complete tool specifications include function descriptions, parameter names, and parameter descriptions. We replace semantic function names with generic identifiers (\texttt{function\_1}, \texttt{function\_2}, etc.), removing the primary semantic cue for tool selection. We then create three different documentation levels:
\begin{enumerate}
    \item \textbf{Anon. function name only}, where we remove everything (function description, parameter names, and parameter descriptions), testing pure behavioral discovery through trial and error with \textbf{no} documentation
    \item \textbf{Anon. function name + Description}, where we remove all parameter names/descriptions while keeping only the function description, testing whether models can infer argument structure from behavioral descriptions alone. 
    \item \textbf{Anon. function name + Parameter names}, where we remove the function description and parameter descriptions while keeping only parameter names, testing discovery of functionality from argument structure without semantic guidance.
\end{enumerate}

\paragraph{Data Collection:} We evaluate on BFCL's executable subset, focusing specifically on the \textbf{simple} and \textbf{multiple} function categories. We restrict our evaluation to these fundamental settings to isolate the model's ability to recover malformed tool descriptions, decoupling this semantic challenge from the combinatorial complexity of parallel execution. This provides a controlled environment to rigorously verify the core mechanism of our approach before scaling to more complex planning scenarios.

\paragraph{Primary Evaluation Metrics:} We measure binary task completion accuracy—whether the model successfully calls the correct tool with proper arguments to satisfy the user query.

\paragraph{Enhanced Evaluation Metrics}
Standard BFCL evaluation relies on a binary success metric (Pass/Fail). To better diagnose \textit{how} agents learn opaque tool behaviors, we introduce two granular metrics that distinguish between semantic understanding (selecting the right tool) and syntactic mastery (calling it correctly).

\textbf{1. Parameter Accuracy.} 
This metric measures the exact correctness of the arguments provided, conditional on the agent selecting the correct tool. If the model chooses the wrong function, the score is 0. When the correct function is chosen, we calculate the percentage of expected parameters that are perfectly recovered. Specifically, it is the ratio of arguments where both the parameter name and the assigned value exactly match the ground truth, divided by the total number of required parameters. This metric strictly penalizes missing arguments or incorrect values, distinguishing agents that ``know'' the tool from those that merely guess the function name.

\textbf{2. AST (Abstract Syntax Tree) Accuracy.} 
AST Accuracy evaluates the structural validity and ``grammar'' of the tool call, independent of whether the values are correct. It is calculated as the average of five components:
\begin{itemize}
    \item \textbf{Format Validity:} Whether the output is parsable as valid JSON or a Python Abstract Syntax Tree (using \texttt{ast.parse}).
    \item \textbf{Structure Validity:} Whether the parsed object contains the standard \texttt{function} and \texttt{args} keys.
    \item \textbf{Type Correctness:} The percentage of parameters where the data type (e.g., string, integer, list) matches the ground truth schema.
    \item \textbf{Schema Compliance:} A strict check ensuring the structure is valid, all types are correct, and no hallucinated parameters exist.
    \item \textbf{Hallucination Check:} Whether the agent generated parameters that do not exist in the tool definition.
\end{itemize}

\subsection{Chess: Learning Strategic Tool Selection Through Experience}
\label{appendix:Chess}

We challenge LLMs to play Chess, but instead of predicting moves directly \jack{cite, theres a paper about this}, models are given access to several undocumented tools that accept current board positions in FEN notation and return move recommendations. Each next move suggestion function has an identical interface, but undocumented behavioral differences. Thus, the agent must discover through gameplay that each implements different strategies. Performance directly reflects the model's ability to document each tool's strengths. %

\paragraph{Task Setup:} Models must maximize playing strength against a series of increasingly hard opponents (Stockfish \jack{cite} configured with ELO ratings 1200, 1800, and 2400) by selecting from available tool sets. %
Each trajectory consists of a max number of moves where the model must select a tool and play against the opponent.

\paragraph{Tool Sets:} We construct two tool sets of increasing complexity: 
\begin{enumerate}
    \item \textbf{Phase specialization} (4 tools) - These tools are engines that work well for specific phases: opening, middlegame, endgame, and late endgame. These phases are defined by the number of pieces on the board: opening phase has at least 28 pieces, middlegame at least 16, endgame at least 10, and late endgame has less than 10 pieces. Each tool plays moves according to a strong engine (Stockfish 2400 ELO) in its own phase but plays randomly otherwise. An optimal agent would learn to document these temporal patterns.
    \item \textbf{Playing Strength gradients} (3 tools) - Tools 1, 2, and 3 are Stockfish configured with ELO ratings 1200, 1800, and 2400 (higher=better); this tests fine-grained discrimination between similar high-quality tools.
\end{enumerate}

\paragraph{Data Collection:} We sample 2000 chess positions from the Lichess database\footnote{\url{https://database.lichess.org}}, which provides hundreds of millions of positions with chess engine evaluations. We split these 2000 positions into training (10\%) and test (90\%) sets, maintaining the same stratified distribution across both game phases and position evaluations to ensure comparable evaluation conditions:
\begin{itemize}
\item \textbf{Game phase} (determined by piece count): opening (25\%), middlegame (40\%), endgame (25\%), late endgame (10\%)
\item \textbf{Position evaluation} (from Lichess engine analysis): equal positions (40\%), slight advantages for white/black (10\% each), winning positions for white/black (8\% each), and crushing/mate positions for each side (6\% each)
\end{itemize}
This stratified sampling ensures models encounter diverse board states that test tool selection across different game scenarios. 

\paragraph{Main Evaluation Metrics:} Since we know the optimal tool call at every turn (the correct phase specialized tool for phase specialization or the highest search depth tool for depth gradients), we simply calculate the accuracy of LLM agent tool calls as our evaluation metric. 

\subsection{Additional Chess Evaluation Metric: Streaming Elo}
\label{sec:chess_elo_details}

To provide a fine-grained measurement of strategic decision quality beyond binary tool-choice accuracy, we implement a \textbf{Streaming Elo} rating system. The Elo rating system is a method for calculating the relative skill levels of players in zero-sum games.

\paragraph{Opponent Pool.} We evaluate the agents against a diverse set of deterministic opponents using the Stockfish engine at varying difficulty levels to represent different tiers of play:
\begin{itemize}
    \item \textbf{Beginner:} Stockfish Level 1 (Approx. Elo 800)
    \item \textbf{Intermediate:} Stockfish Level 5 (Approx. Elo 1600)
    \item \textbf{Master:} Stockfish Level 10 (Approx. Elo 2400)
\end{itemize}

\paragraph{Update Rule.}
The agent starts with a standard baseline rating of $R_0 = 1200$. After each game $i$, the rating is updated based on the result against an opponent with rating $R_{opp}$. We use a K-factor of $K=32$. The expected score $E_i$ and the updated rating $R_{i+1}$ are calculated as follows:

\begin{equation}
    E_i = \frac{1}{1 + 10^{(R_{opp} - R_{i})/400}}
\end{equation}

\begin{equation}
    R_{i+1} = R_i + K \cdot (S_{actual} - E_i)
\end{equation}

where $S_{actual}$ is the game outcome ($1.0$ for a win, $0.5$ for a draw, $0.0$ for a loss).

\paragraph{Experimental constraints and Bootstrapping.}
Due to the high computational cost of running full tool-augmented chess trajectories, we evaluate on a subset of 300 games played against the opponent pool. To prevent infinite loops in drawn or lost positions, any game exceeding \textbf{120 moves} is automatically adjudicated as a draw.

Streaming Elo ratings can be sensitive to the specific chronological order of matches (e.g., facing a string of Master-level opponents early can depress the rating, making recovery difficult). To eliminate this variance and ensure a robust final metric, we employ \textbf{bootstrapping}. We shuffle the sequence of the 300 completed games into \textbf{1,000 random permutations}, calculate the final streaming Elo for each permutation, and report the mean rating across all permutations.

\subsection{BrowseComp Domains: Learning Multi-Tool Coordination for Complex Information Seeking}
\label{appendix:BrowseComp}

Complex question-answering requires discovering not just individual tool capabilities, but how to coordinate multiple tools strategically. BrowseComp Plus \citep{Chen2025BrowseCompPlusAM} provides an ideal testbed for this challenge—human-curated questions that demand synthesizing information from dozens of search queries. Unlike simple retrieval tasks, these questions require models to discover through interaction which tools access which information sources, how to formulate effective queries for each, and how to combine results to build comprehensive answers.

\paragraph{Task Setup:} Models must answer complex, multi-hop questions using search tools with opaque documentation. Each question requires aggregating information from multiple sources—for example, comparing statistics across countries, tracing historical developments, or synthesizing technical specifications. While BrowseComp Plus provides a fixed corpus containing all necessary documents, models receive no documentation about which tools search which subsets or how query syntax varies between tools. They must discover these constraints through experimentation during actual question-answering trajectories.

\paragraph{Tool Sets and Degradation Strategy:} We construct two search environments that test different aspects of tool discovery: (1) \textbf{Domain-specific search} (9 tools) where specialized tools each query distinct document subsets (academic papers, product catalogs, geographical data, news articles), testing discovery of tool coverage boundaries and domain-specific query patterns \jack{we should mention that we do not provide function names}; and (2) \textbf{Mixed search} (10 tools) which combines specialized domain tools with a general tool that searches the entire corpus, testing strategic selection between targeted and broad search approaches. We introduce realistic opacity patterns that mirror production search systems—tools are provided with generic names (\texttt{search\_1}, \texttt{search\_2}) and minimal documentation. Models must discover through interaction: coverage boundaries (which document types each tool can access), query constraints (maximum query length, required syntax, boolean operator support), and ranking behaviors (how each tool prioritizes results by recency, relevance, or popularity). These undocumented behaviors only emerge through varied usage patterns across multiple queries.

\paragraph{Data Collection:} We use BrowseComp Plus's curated question set, which includes 830 complex questions designed to require extensive information gathering. Questions span diverse domains including science, history, geography, and current events. Each question has human-validated answers and requires on average 15-30 search queries when using well-documented tools, making this an ideal benchmark for measuring if models can learn tool capabilities while solving real tasks.

\paragraph{Evaluation Metrics:} We measure both answer accuracy (F1 score against gold answers) and search efficiency (number of queries required). Unlike single-shot benchmarks, we track improvement across questions—does the model become more efficient at using discovered tool capabilities? We also measure cross-question transfer: when models discover a tool searches academic papers while answering a science question, can they apply this knowledge to a history question requiring scholarly sources?

\section{Baselines}
\label{appendix:baselines}
 Following \methodname, for all baselines we use GPT-5. 
 \subsection{Play2Prompt}
 \textbf{Play2Prompt} \citep{Fang2025PLAY2PROMPTZT} improves tool-documentation from self-play followed by self-reflection. It iteratively generates a set of tool usage examples by ``playing'' with the tool, using the responses until it generates valid example tool usages. Using these examples, the documentation is iteratively improved based on the tool use errors observed while using the current documentation.
 \subsection{EasyTool}
 \citep{Yuan2024EASYTOOLEL} which automatically rewrite the tool documentation in two stages. First, it condenses the tool descriptions to eliminate redundant information and focuses only on core functionality. Then, it creates structured functional guidelines with usage scenarios and parameter examples to help LLMs understand when and how to use each tool. EasyTool is limited by its lack of execution of the tools themselves. Furthermore, the descriptions and functional guidelines are beforehand, hence cannot benefit from any knowledge gained as the trajectory rolls out.

\section{\methodname Prompts}
\label{appendix:prompts}

\lstset{
  breaklines=true,
  breakautoindent=false,
  breakindent=0pt
}
\begin{figure}[h]
\centering
\begin{lstlisting}[style=prompt]
You are an expert in composing and exploring functions. You are given a user question and a set of available tools.

You must call at least one tool in response to every user question. There are no exceptions. Refusing to call a tool is not allowed.

If you are confident in a tool's purpose, use it appropriately to address the user's request. If you are unsure what a tool does, make a best guess and try it with plausible parameters to learn how it behaves. It is better to experiment than to fail to respond.

Always format tool calls correctly with all required parameters. You should only return function calls in the tool call sections.
\end{lstlisting}
\caption{BFCL Exploration Prompt}
\label{fig:bfcl_exploration}
\end{figure}

\begin{figure}[h]
\centering
\begin{lstlisting}[style=prompt]
"pre": """You are improving function documentation by analyzing real usage examples. Your goal is to write descriptions that help future users understand exactly what each function does and how to use it correctly.

Current Function Definitions:
{available_functions}

Observed usage examples:
""",
-------------------------------------------------------------------------
"middle": """
Example {example_num}:
User Question: {question}
Function Called: {function_call}
Function Output: {function_output}
""",
-------------------------------------------------------------------------
"post": """
## Important

**If the existing description is already accurate, do not include it in your response.** Only provide updates for functions that actually need improvement. You can tell a description is accurate if in the usage examples the function was called successfully and produced the expected output.

## Instructions

**Only analyze and update functions that were actually called in the examples above.** Do not provide descriptions for functions that weren't used.

For each function that was called:
1. Examine what inputs were provided and what the function returned
2. Compare this with the existing description
3. **Only provide updated descriptions for functions that need improvement**

## Requirements for Updated Descriptions

Each improved description must clearly explain:
- **What the function does** (its purpose and behavior)
- **What inputs it expects** (parameter types, format, requirements)
- **What it returns** (output format and content)
- **How to call the function** (usage pattern or syntax)

Write 1-2 clear, comprehensive sentences that give users everything they need to use the function correctly.

## Format

FUNCTION: [function_name]
DESCRIPTION: [Your improved description here]

## Updated Descriptions (only for functions that need improvement):"""
\end{lstlisting}
\caption{The BFCL reflection prompt. It is split into three parts. A ``pre-prompt", a ``middle-prompt", and a ``post-prompt". We concatenate them together along with the real usage behaviors (ie, the function calls and the outputs).}
\label{fig:bfcl_update}
\end{figure}

\begin{figure}[h]
\centering
\begin{lstlisting}[style=prompt]
You are a chess-playing AI assistant with access to chess analysis tools for evaluating positions and suggesting moves.

<objective>
Play the strongest move in the current position by making exactly ONE tool call to analyze the position.
</objective>

<tool_selection_strategy>
- Make exactly ONE tool call per move decision
- Choose the strongest tool available for the current position
- If tools appear similar in capability or are similarly unknown, consider trying a different tool than recently used ones to diversify your information sources
- When one tool is clearly superior for the position type, use it regardless of history
</tool_selection_strategy>

<exploration_principle>
- Primary goal: Select the strongest tool for each position
- Secondary consideration: If multiple tools seem equally strong or equally unknown, vary your selection based on recent usage history
- This diversification helps avoid potential blind spots from relying on a single tool's perspective
- Never sacrifice move quality for exploration - only explore when tools are genuinely comparable
</exploration_principle>

<decision_framework>
With your single tool call, consider:
- What type of position is this? (tactical, positional, endgame, opening)
- Which tool is strongest for this specific position?
- If multiple tools seem equally strong, which have I used recently?
- Is there a clear best tool, or are several tools comparably suitable?
</decision_framework>

<tool_preamble>
Before making your tool call:
- Explain which tool you're selecting and why it's the strongest choice for this position
- If multiple tools seemed equally viable, briefly note why you selected this one over the others
</tool_preamble>

<quality_checks>
- Select the strongest available tool (or make a reasonable choice among equals)
- Make exactly one tool call
</quality_checks>
\end{lstlisting}
\caption{Chess exploration prompt. We append the board state and the recent tool-calls to this prompt before receiving the next tool-call.}
\label{fig:chess_exploration}
\end{figure}

\begin{figure}[h]
\centering
\begin{lstlisting}[style=prompt]
Analyze chess tool performance across N game trajectories to generate improved tool descriptions that clearly differentiate when to use each tool.

<input>
- Game trajectories with tool calls, moves, and positions
- Board evaluations (positive=White advantage, negative=Black advantage)
- Current tool descriptions
- Side played by agent in each game
</input>

<analysis_requirements>
For each tool:
- Identify consistent patterns in its behavior and performance
- Determine what distinguishes it from other tools
- Provide concrete proof: cite specific trajectories and moves showing these patterns
- Focus on situations where this tool performs differently than others

Evaluation notes:
- Higher eval is better for White, lower eval is better for Black
- IMPORTANT: Always compare tools relatively, not absolutely
- Example for White: Tool A suggesting move to +2 is better than Tool B suggesting +1
- Example for Black: Tool A suggesting move to -3 is better than Tool B suggesting -1
- Critical: Even in losing positions, compare which tool finds the best continuation
  * For White: -5 is much better than -10 (both losing, but one is more resilient)
  * For Black: +10 is much better than +15 (both losing, but one offers more resistance)
- Don't dismiss a tool just because it suggested moves in bad positions - focus on whether it found the BEST move among the alternatives
</analysis_requirements>

<output_per_tool>
**Tool: [name]**

Observed patterns: [Key behaviors identified with specific trajectory evidence]

Distinguishing characteristics: [What makes this tool different from others, with examples]

Updated description:
[Concise description stating when to use this tool relative to others]

Reasoning: [Justification based on trajectory evidence]
</output_per_tool>

<final_output>
After analyzing all tools, provide a decision framework for selecting between tools based on the patterns discovered.
</final_output>

Key: Every claim must reference trajectories. Descriptions must be comparative (tool X better than Y for Z) not absolute.
\end{lstlisting}
\caption{The Chess batch analysis reflection prompt. The chess trajectories are appended to this prompt.}
\label{fig:chess_update}
\end{figure}

\begin{figure}[h]
\centering
\begin{lstlisting}[style=prompt]
You will receive N LLM responses, each analyzing different batches of chess game trajectories. Synthesize these into definitive tool descriptions.

<synthesis_task>
For each tool:
1. Identify patterns that appear across multiple responses
2. Note contradictions between responses
3. Distinguish true patterns from batch-specific noise
4. Look for emergent patterns that no single analysis identified but become visible when viewing all analyses together
5. Create ONE final description based on the most reliable patterns

Critical: 
- A behavior mentioned in only 1-2 responses is likely batch-specific noise
- Focus on patterns that multiple independent analyses discovered
- Also identify meta-patterns: behaviors that emerge from the collective evidence but weren't explicitly stated in any single response
- When responses conflict, examine their evidence strength
- Final descriptions should capture the tool's strengths/weaknesses but NOT explicitly name other tools
</synthesis_task>

<output_format>
**Tool: [name]**

Synthesis reasoning:
[Explain which patterns were most consistent across analyses, what emergent patterns were discovered, how conflicts were resolved, and why certain behaviors were included/excluded in the final description.]

Final description:
[Single definitive description of when to use this tool. Describe its characteristics and optimal use cases WITHOUT referencing other tools by name. Example: "Best for tactical positions requiring deep calculation. Excels at finding forcing sequences and material sacrifices. Tends to be overly aggressive in quiet positions."]
</output_format>
\end{lstlisting}
\caption{The Chess consensus merge reflection prompt. The chess descriptions generated from the previous step are appnded to this prompt.}
\label{fig:chess_merge}
\end{figure}

\begin{figure}[h]
\centering
\begin{lstlisting}[style=prompt]
You are a question-answering AI assistant with access to search tools that return different types of results.

<objective>
Find the correct answer to the question by making strategic tool calls over multiple turns. Each turn, you make exactly ONE tool call and receive its results before deciding your next action.
</objective>

<how_this_works>
- You will be called multiple times for the same question
- Each time, you'll see the full history of your previous tool calls and their results
- Each turn, make exactly ONE tool call to gather more information
- Use what you've learned from previous tool calls to inform your next choice
- Once you have enough information, provide your final answer
</how_this_works>

<tool_selection_strategy>
- Make exactly ONE tool call per turn
- Choose the strongest tool available for the current information needs
- Review what you've already learned from previous tool calls
- If tools appear similar in capability or you're uncertain about what they return, consider trying a different tool than recently used ones to diversify your information sources
- When one tool is clearly superior for the remaining information needs, use it regardless of history
</tool_selection_strategy>

<exploration_principle>
- Primary goal: Select the strongest tool for your current information gap
- Secondary consideration: If multiple tools seem equally strong or you're uncertain about their outputs, vary your selection based on what you've already tried
- This diversification helps avoid potential blind spots from relying on a single tool's perspective
- Never sacrifice answer quality for exploration - only explore when tools are genuinely comparable or unknown
- Learn from previous tool results: if a tool gave poor results before, consider alternatives
</exploration_principle>

<decision_framework>
Each turn, consider:
- What information do I still need to answer this question?
- What have I learned from previous tool calls?
- What type of question is this? (factual, current events, historical, technical, domain-specific)
- Which tool is strongest for filling my current information gap?
- If multiple tools seem equally strong or I'm uncertain about them, which have I used already?
- Do I have enough information to answer, or should I make another tool call?
</decision_framework>

\end{lstlisting}
\caption{BrowseComp Domains exploration prompt, part 1. We append the context, including the recent tool-calls and results, to this prompt before receiving the next tool-call.}
\label{fig:bc_exploration}
\end{figure}

\begin{figure}[h]
\centering
\begin{lstlisting}[style=prompt]
<tool_preamble>
Before making your tool call:
- Review what you've learned from previous tool calls (if any)
- Explain which tool you're selecting and why it's the strongest choice for your current information needs
- If you're uncertain about what a tool returns, acknowledge this uncertainty
- If multiple tools seemed equally viable or unknown, briefly note why you selected this one over the others
</tool_preamble>

<quality_checks>
- Review previous tool calls and their results
- Select the strongest available tool for your current needs (or make a reasonable choice among equals/unknowns)
- Make exactly one tool call per turn
- Use accumulated tool results across turns to formulate your answer
</quality_checks>
\end{lstlisting}
\caption{BrowseComp Domains exploration prompt, part 2. We append the context, including the recent tool-calls and results, to this prompt before receiving the next tool-call.}
\label{fig:bc_exploration2}
\end{figure}

\begin{figure}[h]
\centering
\begin{lstlisting}[style=prompt]
Analyze search tool performance across N question-answering trajectories to generate improved tool descriptions that clearly differentiate when to use each tool.

<input>
- Question-answering trajectories with tool calls and results
- Search results returned by each tool (content may vary by tool)
- Whether the final answer was correct or incorrect
- Current tool descriptions
</input>

<analysis_requirements>
For each tool:
- Identify consistent patterns in the type and quality of results it returns
- Determine what distinguishes it from other tools
- Provide concrete proof: cite specific trajectories and queries showing these patterns
- Focus on situations where this tool performs differently than others

Evaluation notes:
- IMPORTANT: Compare tools relatively, not absolutely
- A tool is effective if it helps the agent reach the correct answer
- Consider both successful and unsuccessful trajectories
- Focus on: result relevance, information completeness, and query type suitability
- Don't dismiss a tool just because it was used in failed trajectories - focus on whether it provided useful information compared to alternatives
</analysis_requirements>

<output_per_tool>
**Tool: [name]**

Observed patterns: [Key behaviors identified with specific trajectory evidence]

Distinguishing characteristics: [What makes this tool different from others, with examples]

Updated description:
[Concise description stating when to use this tool relative to others]

Reasoning: [Justification based on trajectory evidence]
</output_per_tool>

<final_output>
After analyzing all tools, provide a decision framework for selecting between tools based on the patterns discovered.
</final_output>

Key: Every claim must reference trajectories. Descriptions must be comparative (tool X better than Y for Z) not absolute.
\end{lstlisting}
\caption{The BrowseComp domains batch analysis reflection prompt. The trajectories are appended to this prompt.}
\label{fig:bc_update}
\end{figure}

\begin{figure}[h]
\centering
\begin{lstlisting}[style=prompt]
You will receive N LLM responses, each analyzing different batches of question-answering trajectories. Synthesize these into definitive tool descriptions.

<synthesis_task>
For each tool:
1. Identify patterns that appear across multiple responses
2. Note contradictions between responses
3. Distinguish true patterns from batch-specific noise
4. Look for emergent patterns that no single analysis identified but become visible when viewing all analyses together
5. Create ONE final description based on the most reliable patterns

Critical:
- A behavior mentioned in only 1-2 responses is likely batch-specific noise
- Focus on patterns that multiple independent analyses discovered
- Also identify meta-patterns: behaviors that emerge from the collective evidence but weren't explicitly stated in any single response
- When responses conflict, examine their evidence strength
- Final descriptions should capture the tool's strengths/weaknesses but NOT explicitly name other tools
</synthesis_task>

<output_format>
**Tool: [name]**

Synthesis reasoning:
[Explain which patterns were most consistent across analyses, what emergent patterns were discovered, how conflicts were resolved, and why certain behaviors were included/excluded in the final description.]

Final description:
[Single definitive description of when to use this tool. Describe its characteristics and optimal use cases WITHOUT referencing other tools by name. Example: "Best for queries requiring recent information or real-time data. Returns comprehensive results with detailed snippets. May be less effective for historical or archival content."]
</output_format>
\end{lstlisting}
\caption{The BrowseComp domains consensus merge reflection prompt. The descriptions generated from the previous step are appended to this prompt.}
\label{fig:bc_merge}
\end{figure}
 
\end{document}